\documentclass{article} %
\usepackage{iclr2024_conference,times}

\usepackage{amsmath,amsfonts,bm}

\def\eqref#1{equation~\ref{#1}}

\def\1{\bm{1}}

\DeclareMathAlphabet{\mathsfit}{\encodingdefault}{\sfdefault}{m}{sl}
\SetMathAlphabet{\mathsfit}{bold}{\encodingdefault}{\sfdefault}{bx}{n}

\usepackage{adjustbox}
\usepackage{amsfonts}
\usepackage{amsmath}
\usepackage{amssymb}
\usepackage{bm}
\usepackage{booktabs}
\usepackage{colortbl}
\usepackage{comment}
\usepackage{dsfont}
\usepackage{enumitem}
\usepackage{graphicx} 
\usepackage{hyperref}
\usepackage{inconsolata}
\usepackage{mathtools}
\usepackage{microtype}
\usepackage{multicol}
\usepackage{multirow}
\usepackage{nicefrac}
\usepackage{lscape}
\usepackage{scalerel}
\usepackage{soul}
\usepackage{subcaption}
\usepackage{tabularx}
\usepackage[most]{tcolorbox}
\usepackage{url}
\usepackage{xcolor}
\usepackage{xspace}
\usepackage[T1]{fontenc}
\usepackage{wrapfig}

\newcommand\blankfootnote[1]{%
  \let\thefootnote\relax\footnotetext{#1}%
  \let\thefootnote\svthefootnote%
}

\newcommand{\ssc}[1]{{\small \sc #1}\xspace}
\newcommand{\bssc}[1]{{\small \sc \textbf{#1}}\xspace}
\newcommand{\issc}[1]{{\small \sc \textit{#1}}\xspace}
\newcommand{\fssc}[1]{{\scriptsize \sc #1}\xspace}

\newcommand{\freshqa}{{\ssc{FreshQA}}\xspace}
\newcommand{\freshprompt}{{\ssc{FreshPrompt}}\xspace}
\newcommand{\fresheval}{{\ssc{FreshEval}}\xspace}
\newcommand{\seal}{{\ssc{FreshPrompt}}\xspace}
\newcommand{\llm}{{\ssc{LLM}}\xspace}
\newcommand{\llms}{{\ssc{LLMs}}\xspace}
\newcommand{\strict}{{\ssc{Strict}}\xspace}
\newcommand{\relaxed}{{\ssc{Relaxed}}\xspace}
\newcommand{\smallurl}[1]{\begin{small}\url{#1}\end{small}}
\newcommand{\chatgpt}{{\ssc{ChatGPT}}\xspace}
\newcommand{\gptfour}{{\ssc{GPT-4}}\xspace}
\newcommand{\gptthreefive}{{\ssc{GPT-3.5}}\xspace}
\newcommand{\codex}{{\ssc{Codex}}\xspace}
\newcommand{\tfive}{{\ssc{T5}}\xspace}
\newcommand{\palm}{{\ssc{PaLM}}\xspace}
\newcommand{\flanpalm}{{\ssc{Flan-PaLM}}\xspace}

\newcommand{\fewshotprompt}{{\ssc{Few-shot}}\xspace}
\newcommand{\cotprompt}{{\ssc{CoT}}\xspace}

\newcommand{\selfask}{{\ssc{Self-Ask}}\xspace}
\newcommand{\google}{{\ssc{Google Search}}\xspace}

\definecolor{myblue}{HTML}{2B79B0}
\definecolor{myorange}{HTML}{FB7F36}
\definecolor{mygreen}{HTML}{389E3B}
\definecolor{myred}{HTML}{D22D35}

\title{\centering \textsc{FreshLLMs}: \\ Refreshing Large Language Models \\ with Search Engine Augmentation}
\newcommand{\hspaceone}{\hspace{4.0ex}}
\newcommand{\hspacetwo}{\hspace{2.0ex}}

\author{\centerline{Tu Vu$^{1}$ \hspaceone Mohit Iyyer$^{2}$ \hspaceone Xuezhi Wang$^{1}$ \hspaceone Noah Constant$^{1}$ \hspaceone Jerry Wei$^{1}$} \\[1.0ex] \bf \centerline{Jason Wei$^{3}$\thanks{Work done while at Google.}  \hspacetwo Chris Tar$^{1}$ \hspacetwo Yun-Hsuan Sung$^{1}$ \hspacetwo Denny Zhou$^{1}$ \hspacetwo Quoc Le$^{1}$ \hspacetwo Thang Luong$^{1}$} \\[2.0ex]
\centerline{Google$^1$ \hspaceone University of Massachusetts Amherst$^2$ \hspaceone OpenAI$^3$} \\
\centerline{\texttt{freshllms@google.com}} \vspace{-2.0ex}\\}
\vspace{-2mm}

\iclrfinalcopy %

\begin{document}

\maketitle

\begin{abstract}
\label{section:abstract} 
Most large language models (\llms) are trained once and never updated; thus, they lack the ability to dynamically adapt to our ever-changing world. In this work, we perform a detailed study of the factuality of \llm-generated text in the context of answering questions that test current world knowledge. Specifically, we introduce \ssc{FreshQA}, a novel dynamic \ssc{QA} benchmark encompassing a diverse range of question and answer types, including questions that require \textit{fast-changing} world knowledge as well as questions with \textit{false premises} that need to be debunked. We benchmark a diverse array of both closed and open-source \llms under a two-mode evaluation procedure that allows us to measure both correctness and hallucination. 
Through human evaluations involving more than \ssc{50K} judgments, we shed light on limitations of these models and demonstrate significant room for improvement: for instance, all models (regardless of model size) struggle on questions that involve fast-changing knowledge and false premises. Motivated by these results, we present \freshprompt, a simple few-shot prompting method that substantially boosts the performance of an \llm on \freshqa by incorporating relevant and up-to-date information retrieved from a search engine into the prompt.
Our experiments show that \freshprompt\ outperforms both competing search engine-augmented prompting methods such as \ssc{Self-Ask}~\citep{OPress22} as well as commercial systems such as \ssc{Perplexity.AI}.\footnote{\smallurl{https://www.perplexity.ai}} Further analysis of \freshprompt\ reveals that both the number of retrieved evidences and their order play a key role in influencing the correctness of \llm-generated answers. Additionally, instructing the \llm\ to generate concise and direct answers helps reduce hallucination compared to encouraging more verbose answers. To facilitate future work, we release \freshqa\ at \href{https://github.com/freshllms/freshqa}{\textcolor{mygreen}{\begin{small}\texttt{github.com/freshllms/freshqa}\end{small}}} and commit to updating it at regular intervals.
\end{abstract}

\section{Introduction}
\label{section:introduction}
Recent large language models (\llms) such as \ssc{Bard} and \ssc{ChatGPT/GPT-4}\footnote{\smallurl{https://bard.google.com}, \smallurl{https://chat.openai.com}} are designed to be versatile open-domain \textit{chatbots} that can engage in multi-turn conversations on diverse subjects.
Despite their impressive capabilities, these \llms often ``hallucinate'' plausible but factually incorrect information ~\citep{JMaynez20,NLiu23a}, which reduces the trustworthiness of their responses, especially in settings where accurate and up-to-date information is critical. This behavior can be partially attributed to the presence of outdated knowledge encoded in their parameters. While additional training using human feedback~\citep{LOuyang22} or knowledge-enhanced tasks can mitigate this issue, it is not easily scalable for real-time knowledge updates (e.g., stock price of a company). %
In-context learning~\citep{TBrown20} is an appealing alternative in which real-time knowledge can be injected into an \llm's prompt for conditioning generation. While recent work has begun to explore augmenting \llms\ with web search results~\citep{ALazaridou22,OPress22}, it is unclear how to take full advantage of search engine outputs to increase \llm\ factuality.

\begin{figure}[t]
\centering
\includegraphics[width=\textwidth]{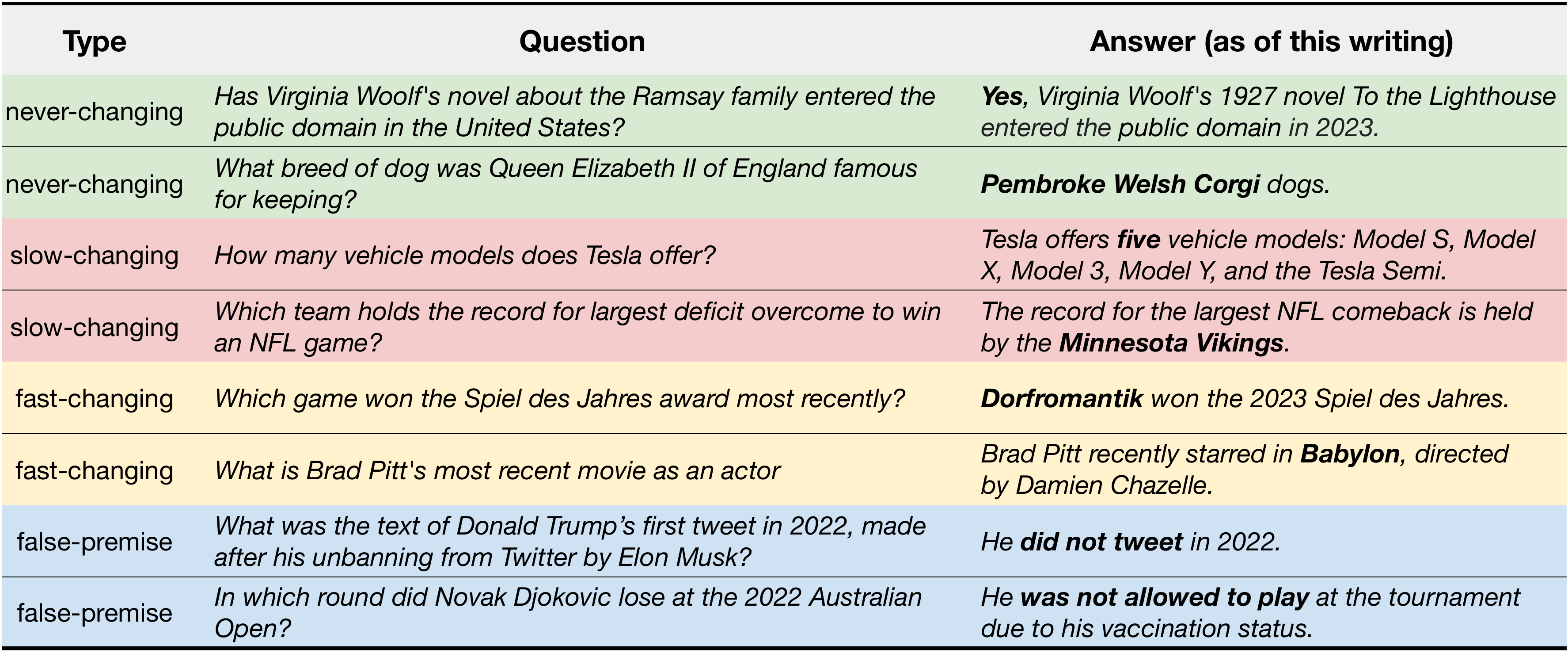}
\caption{\freshqa exemplars. Our questions are broadly divided into \textit{four} main categories based on the nature of the answer: \textit{\textcolor{mygreen}{never-changing}}, in which the answer almost never changes; \textit{\textcolor{mygreen}{slow-changing}}, in which the answer typically changes over the course of several years; \textit{\textcolor{mygreen}{fast-changing}}, in which the answer typically changes within a year or less; and \textit{\textcolor{mygreen}{false-premise}}, which includes questions whose premises are factually incorrect and thus have to be rebutted.
}
\label{figure:freshqa_examples}
\vspace{-3mm}
\end{figure}
In this work, we collect a novel \ssc{QA} benchmark, dubbed \freshqa, to evaluate the factuality of existing \llms. \freshqa consists of 600 natural questions that are broadly divided into the \textit{four} main categories shown in Figure~\ref{figure:freshqa_examples}.
\freshqa's questions span a diverse set of topics with diverse difficulty levels (requiring single-hop and multi-hop reasoning), and require a model to ``understand'' the world's up-to-date knowledge to be able to answer correctly.  Additionally, \freshqa is dynamic in nature: some of the ground-truth answers may change over time, and a question classified under a specific category may undergo reclassification at some later point in time (e.g., the current \textit{false-premise} question \textit{``How long has Elon Musk been married to his current spouse?''} will fall into the \textit{fast-changing} category if Elon Musk gets married again in the future).

We benchmark how well different \llms perform on \freshqa  by prompting them with questions and optionally a few question-answer demonstrations  %
and then sampling a response. Then, we conduct an extensive human evaluation of the factual accuracy of the models' responses, consisting of more than \ssc{50K} judgements. We evaluate each response in a two-mode evaluation procedure: \relaxed, which measures only whether the main answer is correct; and \strict, which measures whether all of the claims in the response are factual and up-to-date (i.e., no hallucination). Our study sheds light on the factuality of old and new \llms and reveals different model behaviors across question types. Unsurprisingly, there are flat scaling curves on questions that involve fast-changing knowledge: simply increasing the model size does not lead to reliable performance gains. We also observe similar trends on false-premise questions, though several \llms are able to debunk a false-premise question if explicitly asked \textit{``Please check if the question contains a valid premise before answering''}. Overall, \freshqa is challenging for current \llms and leaves ample room for improvement.

Motivated by these findings, we further investigate how to effectively improve \llms' factuality by grounding their responses to accurate and up-to-date information from search engines. Given the rapid development of ever larger \llms and the ever-changing nature of knowledge, we explore in-context learning approaches that allow an \llm to attend over knowledge provided at inference time through its prompt. %
We develop \freshprompt, a simple yet effective method that, for a given question, takes full advantage of a search engine by extracting all up-to-date and relevant information (including knowledge from relevant questions that search users also ask) and uses few-shot in-context learning to teach a model to reason over retrieved evidences and figure out the right answer.
We show that \freshprompt significantly boosts \llms's factuality: for example, our best \gptfour + \freshprompt variant yields an improvement of 32.6\% and 49.0\% accuracy over the vanilla \ssc{GPT-4} on \freshqa under \relaxed and \strict, respectively. Since our method requires no additional training, it is flexible and applicable to a variety of scenarios.

Taken together, our key contributions include:
\begin{itemize}
    \item We introduce a novel dynamic \ssc{QA} benchmark, \freshqa, which features a diverse set of question and answer types, including questions whose answers may change over time and questions whose premises are factually incorrect. We make our dataset freely available and commit to updating the ground-truth answers at a regular schedule to encourage exploration of methods to improve \llms' factuality.
    
    \item We benchmark a wide range of both closed and open-source \llms on our dataset. Through an extensive and rigorous human evaluation study, we shed light on limitations of current \llms: they struggle on fast-changing, false-premise, and multi-hop questions, and our two-mode evaluation captures increased hallucinations produced by  techniques such as chain-of-thought prompting~\citep{JWei22}.
    
    \item We present \seal, a simple in-context learning method that can substantially boost an \llm's factuality compared to competing search-augmented approaches by effectively incorporating factual and up-to-date information from a search engine into the model's prompt. Furthermore, we perform a series of sensitivity and ablation analyses to better understand  what facets of \seal contribute to its success.
\end{itemize}

\section{FreshQA}
 
In this section, we address the growing need to assess \llm\ factuality by curating a novel \ssc{QA} benchmark, \freshqa, with 600 questions that cover a wide spectrum of question and answer types.

\subsection{Data collection}
We collected \freshqa by recruiting both \ssc{NLP} researchers (including the authors and their colleagues) and online freelancers\footnote{We use \fssc{Upwork} (\smallurl{https://www.upwork.com}) with a compensation rate of \$2 per example.} to write questions of varying difficulty levels and topics whose answers may  change based on new developments in the world. The annotators were shown a few exemplars of the four broad types of questions defined in Figure~\ref{figure:freshqa_examples}. 
Within each of these four categories, we ask annotators to write questions at two different difficulty levels: \textit{one-hop}, where the question explicitly mentions all of the relevant information needed to answer it, and thus no additional reasoning is required (e.g., \textit{``Who is the CEO of Twitter''}); and \textit{multi-hop}, where the question requires one or more additional steps of reasoning in order to gather all of the relevant information needed to answer it (e.g., \textit{``What is the total height of the tallest building in the world?''}). Annotators were encouraged to write questions that involve \textit{fresh} knowledge (knowledge that has changed recently or new events) and appear \textit{natural} (i.e.,  plausible for a real person to type into a search engine). For false-premise questions, we requested a brief explanation elucidating why the question is flawed.\footnote{Additionally, the annotators were asked to include the year the answer to the question last changed and an \fssc{URL} to a reputable website that supports the answer.}

\vspace{-1em}
\paragraph{Quality control:} Upon obtaining the initial dataset, we conducted multiple thorough data cleaning and quality assessments. This involved manual review of each example to ensure well-formed questions, removal of duplicates and invalid questions (e.g., too easy or controversial), and verification of answers and supporting evidence \ssc{URLs}. We also manually collected supplementary valid answers for each question (e.g., different names of the same person, different date formats, etc.). To facilitate future answer updates, we excluded questions whose answers  are likely to change more frequently than once per week, and additionally incorporated the expected next review date for each question.

\vspace{-1em}
\paragraph{Data size and split:}
The resulting dataset is divided into a \textit{test} set consisting of 125 questions for each of the four broad question types (500 total examples) and a \textit{development} set comprising 25 questions for each question type (100 total examples), sampled randomly within types. Additionally, 15 examples spanning different question types were extracted for \textit{demonstration} purposes (i.e., for use in few-shot in-context learning), and the remaining data was discarded. The development set is reserved for future studies and not used in this paper.\footnote{Although our test set is currently balanced across question types, the distribution may change over time due to reclassification of questions from one category to another.}

\vspace{-1em}
\paragraph{\bssc{FreshQA} requires regular updates:} Our dataset has time sensitivity since the ground-truth answers may change with new developments in the world. As such, we commit to updating the dataset regularly
and encourage researchers to evaluate on the latest version of the dataset, as close to the release date of the updated dataset as possible.

\subsection{Evaluation}
All model responses were evaluated by the authors in a two-mode evaluation procedure: \relaxed, which focuses solely on evaluating the correctness of the primary answer; and \strict, which additionally examines whether \emph{all} of the facts in the answer are accurate (i.e., no hallucination).  Overall, our setup provides both ends of the spectrum for evaluating factuality (the difference between a model's strict and relaxed performance provides a way to measure hallucination), offering a more comprehensive and nuanced understanding of their performance.

\vspace{-1em}
\paragraph{Evaluation protocol:} In both evaluation modes, we credit a model's response only if it provides a confident and definitive answer, or the correct answer can be obviously inferred from the response. The primary or final answer when standing alone must be accurate. Any additional information that is provided must not contradict the primary answer or reshape one's perception of it. For false-premise questions, the model must point out the presence of a false premise to receive credit. For answers that involve names of entities (e.g., people), complete names or commonly recognized names are expected. Regarding numerical answers, approximate numbers are generally not accepted unless explicitly included in the ground-truth answers. Under \relaxed, we accept ill-formed responses (including those in a non-English language), as well as hallucinated or outdated information  that does not significantly impact the primary answer. Under \strict, however, a response that contains any hallucination, no matter how minor, will not receive credit. Furthermore, we accept a response in \strict\ when the model indicates that the information might be outdated (e.g., {\textit{``As of my knowledge cutoff date in September 2021''}}) \emph{only} if it is evident that the knowledge has not changed.\footnote{Note that even without access to real-time data, a model may still provide accurate answers to certain questions involving current information, potentially through random guesses %
or by leveraging past valid responses (e.g., for the question \textit{``Which drama series won the most recent Primetime Emmy Award for Outstanding Drama Series?''}, while \textit{``Succession''} won the award most recently (as of this writing), it was also the winner in 2020, so a model trained in 2021 could potentially provide the correct answer).} Figure~\ref{figure:freshqa_eval} in Appendix~\ref{appendix:evaluation_protocol} shows specific examples of each evaluation criteria.
\vspace{-1em}
\paragraph{Inter-rater agreement and automatic evaluation:} Two authors independently evaluated a subset of 100 answers in both modes and had an agreement of 99\% for \relaxed\ and 96\% for \strict, showing that the protocol is reliable for comparing different \llms. \textcolor{black}{Additionally, to facilitate future evaluations, we develop \fresheval, a simple automatic metric that uses few-shot in-context learning to teach an \llm to judge model responses, achieving an average agreement of  96.5\% with human evaluations for \relaxed\ and 96\% for \strict. See Appendix~\ref{appendix:inter_rater_agreement} for details.}

\section{Pre-trained LLMs struggle on FreshQA}
\label{section:ood_llms}
We use \freshqa\ to benchmark  \llms that do not have access to real-time data or the ability to browse the Internet for current information.\footnote{With the exception of \fssc{ChatGPT} and \fssc{GPT-4}, which have access to the current date. Note that the latest versions of these models can now browse the Internet.} While all \llms (regardless of size) predictably struggle on questions requiring up-to-date knowledge, they also underperform on  false premise questions. 
In our experiments, we simply feed individual questions as prompts into each model and decode the model's predictions using a temperature of 0 without fine-tuning (see Appendix~\ref{appendix:ood_llms_details} for more details).%
\vspace{-1em}
\paragraph{Baselines:}
We experiment with a series of models varying in size from \ssc{770M} to \ssc{540B} parameters, including basic pre-trained models such as \bssc{T5}~\citep{CRaffel20,BLester21}, \bssc{PaLM} and \bssc{PaLMChilla}~\citep{AChowdhery22}, optionally using \bssc{Few-shot} prompting~\citep{TBrown20} and Chain-of-Thought~\citep[\bssc{CoT},][]{JWei22};\footnote{As we are interested in exploring how these methods perform without being specifically designed for \fssc{FreshQA}, we use the 5-shot demonstrations for \fssc{TriviaQA}~\citep{MJoshi17} used in~\citet{ZSun23}.} instruction-tuned models including \bssc{FLAN-T5} and \bssc{FLAN-PaLM}~\citep{HWChung22,SLongpre23}, and \ssc{OpenAI}'s \bssc{GPT-3.5}~\citep{LOuyang22}, \bssc{Codex}~\citep{MChen21}, \bssc{ChatGPT},  and \bssc{GPT-4}~\citep{OpenAI23}.

\begin{figure*}[t!]
\centering
\includegraphics[width=\textwidth]{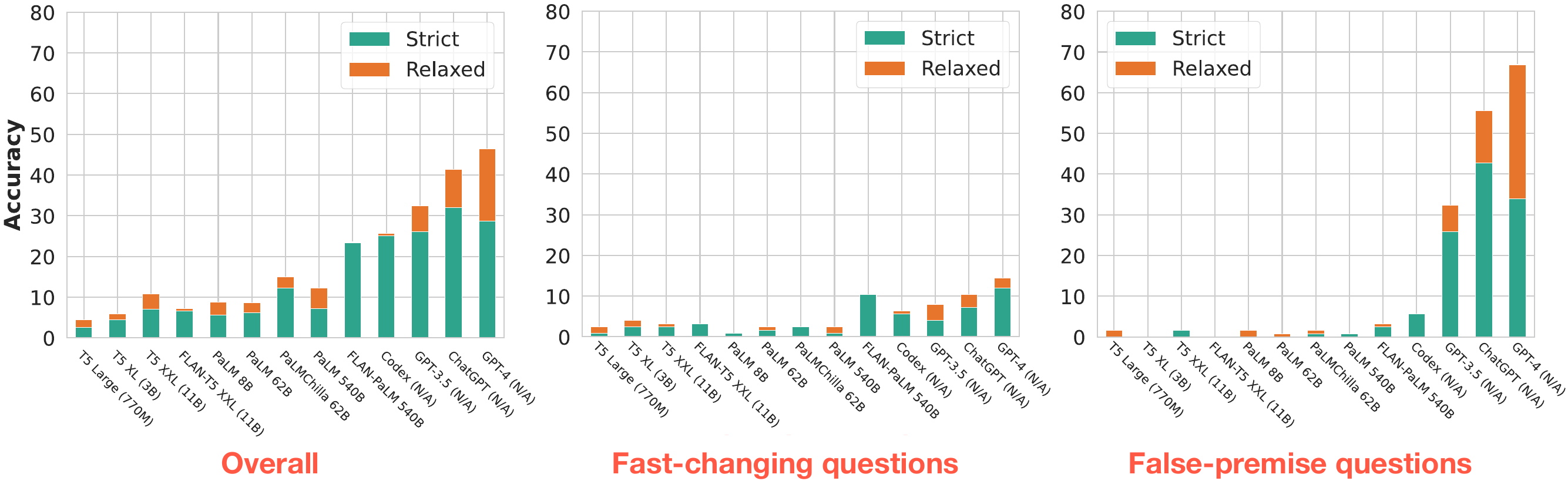}
\caption{Accuracy of different \llms on \freshqa under \relaxed and \strict (no hallucination) evaluations. \hl{Models benchmarked on the same date of April 26, 2023}. \emph{All} models (regardless of model size) struggle on questions that involve \textit{fast-changing} knowledge and \textit{false premises}.}
\label{figure:freshqa_ood_llms}
\vspace{-4mm}
\end{figure*}

\subsection{Results and Discussion}
\label{subsection:ood_llms_results}

\paragraph{\bssc{FreshQA} presents a challenge for \bssc{LLMs}:}
We visualize the accuracy of different \llms on \freshqa in both evaluation modes in Figure~\ref{figure:freshqa_ood_llms}.\footnote{Table~\ref{table:freshqa_strict_ood_llms} and Table~\ref{table:freshqa_relaxed_ood_llms} in Appendix~\ref{appendix:ood_llms_results} contain concrete numbers under \fssc{Strict} and \fssc{Relaxed}, respectively.} 
A first obvious takeaway is that all models struggle on \freshqa: overall accuracy ranges from 0.8\% to 32.0\% under \strict, and 0.8\% to 46.4\% under \relaxed. Switching from \relaxed to \strict results in a marked decrease in accuracy for \chatgpt and \gptfour. This is mainly due to the lack of access to up-to-date information, as they produce ``outdated'' answers (which often start with the prefix `\textit{`As of my knowledge cutoff date in September 2021}''), and in many cases, ``refuse'' to provide an answer (e.g., \textit{``As an \issc{AI} language model, I cannot provide real-time information.''}). %
Similarly, the accuracy of \palm (across model sizes) drops significantly under \strict. Much of this drop is due to artifacts such as conversation-like responses with unexpected special tokens (e.g., the end-of-turn \texttt{[eot]}), and hallucination. In contrast, \ssc{Flan-PaLM} and \ssc{Codex} exhibit minimal hallucination due to their concise and direct answers. 

\vspace{-1em}
\paragraph{\bssc{LLMs} struggle with questions about current information:} 
The lack of up-to-date parametric knowledge results in dramatically degraded accuracies across models on questions involving fast-changing or recent knowledge. \gptfour generally obtains the highest accuracy on these questions, with the exception of questions about recent knowledge (i.e., since 2022) under \strict where it underperforms \ssc{Flan-PaLM} and \ssc{Codex}, but it never exceeds 15\% across both evaluation modes. 
Our evaluation confirms that \chatgpt and \gptfour have been exposed to data containing information beyond their knowledge cutoff date (Appendix~\ref{appendix:chatgpt_gpt4_recent_knowledge}). Additionally, \gptfour is more reluctant to answer fast-changing questions (refusing to answer 60\% of the time) compared to \chatgpt (16\%).

\vspace{-1em}
\paragraph{Questions with false premises pose a hurdle for \bssc{LLMs}:}

All models struggle on questions with false premises, and using larger models does not increase accuracy for \tfive and \palm (``flat scaling''), with performance within the range of 0.0\% to 1.6\%.  \gptthreefive, \chatgpt, and \gptfour demonstrate much superior accuracies to all other models, achieving accuracies between 25.8\% to 42.7\% under \strict and 32.3\% to 66.9\% under \relaxed. \chatgpt performs the best under \strict (42.7\%) while \gptfour is the most accurate model under \relaxed (66.9\%), with an impressive accuracy of 83.9\% on questions about knowledge before 2022. These results suggest that \ssc{OpenAI}'s models are likely trained to cope with false-premise questions.

\vspace{-1em}
\paragraph{\bssc{CoT} increases hallucination:} Overall, \fewshotprompt\ and \cotprompt\ prompting are beneficial for large models %
and sometimes advantageous for moderately-sized models %
on questions with valid premises, especially on questions about never-changing or old knowledge. Under \strict, \fewshotprompt and \cotprompt yields +36.1\% and +26.9\% respective accuracy improvement over zero-shot prompting with \palm\xspace\ssc{540B} on questions involving knowledge before 2022 (+21.9\% and +29.7\% under \relaxed). 
\cotprompt largely demonstrates superior performance compared to \fewshotprompt under \relaxed %
, whereas \fewshotprompt obtains better results under \strict, as \cotprompt introduces more room for hallucination. %

\vspace{-1em}
\paragraph{Multi-hop reasoning is challenging for several models:}%
\tfive\xspace\ssc{Large} and \ssc{XL} are incapable of dealing with multi-hop questions, while \flanpalm\xspace\ssc{540B}, \codex, and \gptthreefive suffer the most when switching from one-hop to multi-hop  questions. \gptfour remains stable across these two types of questions (with a difference of less than 2\% in accuracy across settings). See Appendix~\ref{appendix:ood_llms_results} for details.

\section{Prompting Search Engine-Augmented Language Models}
\label{section:utd_llms}
The low accuracies reported in the previous section are largely unsurprising, as none of the models we evaluated had access to real-time information. In this section, we evaluate the impact of \emph{search engine augmentation} to \llms on \freshqa. We  present \freshprompt, a simple few-shot prompting method that substantially boosts \freshqa performance of an \llm by incorporating relevant and up-to-date information retrieved from a search engine (\google) into the prompt.
\subsection{FreshPrompt}
Our \seal method leverages a text prompt to (1) introduce contextually relevant and up-to-date information (including answers to relevant questions) from a search engine to a pre-trained \llm, and (2) teach the model to reason over retrieved evidences. 
\begin{figure}[t!]
\centering
\includegraphics[width=0.8\textwidth]{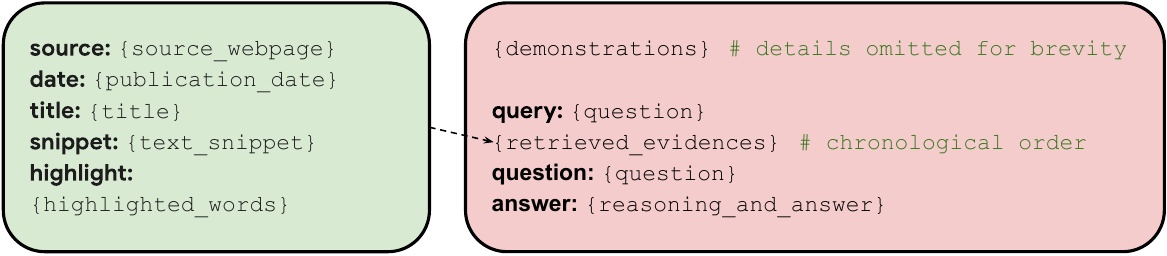}
\caption{\freshprompt's format. We cast all retrieved evidences into a unified format with useful information, including source webpage, date, title, text snippet, and highlighted words (left). Few-shot demonstrations are provided at the beginning of the prompt. Each demonstration shows the model an example question and a list of retrieved evidences for the question, followed by some reasoning over the evidences to figure out the most relevant and up-to-date answer (right).}
\label{figure:freshprompt_format}
\vspace{-3mm}
\end{figure}
More specifically, given a \textit{question} $\bm{q}$, we first use \textit{$\bm{q}$} verbatim to query a search engine, in our case \ssc{Google Search}.\footnote{We scrape the results from \fssc{Google Search} using \fssc{SerpApi} (\smallurl{https://serpapi.com}).} We retrieve all of the search results, including the \textit{answer box}, \textit{organic results}, and other useful information, such as the \textit{knowledge graph}, \textit{questions and answers} from crowdsourced \ssc{QA} platforms, and \textit{related questions} that search users also ask (see Figure~\ref{figure:search_results} in Appendix~\ref{appendix:search_results}). For each of these results, we extract the associated \textit{text snippet} $\bm{x}$ along with other information, such as \textit{source} $\bm{s}$ (e.g., \ssc{Wikipedia}), \textit{date} $\bm{d}$, \textit{title} $\bm{t}$, \textit{highlighted words} $\bm{h}$, and then create a list of $\bm{k}$ retrieved evidences $\bm{E = \{(s, d, t, x, h)\}}$. These evidences are then cast into a common format (Figure~\ref{figure:freshprompt_format}, left) and used to condition the model through in-context learning. 
To encourage the model to focus on more recent evidences, in line with recent findings~\citep{NLiu23b}, we sort the evidences   $\bm{E}$ in the prompt from oldest to newest. 

To help the model to ``understand'' the task and the desired output, we provide  few-shot demonstrations of input-output exemplars at the beginning of the input prompt. Each demonstration shows the model an example question and a list of retrieved evidences for the question, followed by a chain-of-thought reasoning over the evidences to figure out the most relevant and up-to-date answer (Figure~\ref{figure:freshprompt_format}, right).
Although we include a few exemplars of questions with false premises in the demonstrations, we also experiment with an explicit false premise check in the prompt: \textit{``Please check if the question contains a valid premise before answering''}. Figure~\ref{figure:freshprompt_realistic_prompt} in Appendix~\ref{appendix:freshprompt_realistic_prompt} shows a realistic prompt.

\subsection{Experiment setup} 

We closely follow the setup in Section~\ref{section:ood_llms} except in cases where we lack control over the model's decoding via an \ssc{API} (e.g., \ssc{Perplexity.AI}). Some of the models we evaluate can potentially change over time, which presents a challenge to the reproducibility of our evaluation results; \hl{thus, we evaluate all models on the same date of April 26, 2023}.
In addition to \gptthreefive and \gptfour, we evaluate 
\textbf{\bssc{Google Search}} by simply querying \ssc{Google Search} and using the answer in the answer box (if any) or the text snippet of the top-1 search result; 
\textbf{\bssc{Perplexity.AI (PPLX.AI)}}, an answer engine that combines an \llm and a search engine to generate useful responses to users' queries;\footnote{\url{https://www.perplexity.ai}. At the time of evaluation, {\fssc{PPLX.AI}} was a combination of \fssc{GPT-3.5} and \fssc{Bing Search}, and was able to provide both concise and detailed answers. We evaluated its concise answers.} and \textbf{{\bssc{Self-Ask}}}~\citep{OPress22}, a method that uses few-shot in-context learning to teach an \llm to decompose each question into simpler sub-questions that are answered via \google.\footnote{We use the few-shot prompt provided by \fssc{Self-Ask}'s authors and apply it to both \fssc{GPT-3.5} and \fssc{GPT-4}. For simplicity, we evaluate solely the final answer from \fssc{Self-Ask}, disregarding intermediate answers.}

\noindent\textbf{{\bssc{FreshPrompt} setup:}} We apply \freshprompt\ to both \gptthreefive and \gptfour by sequentially incorporating the following retrieved evidences into the input prompt: $\bm{o}$ organic search results, $\bm{r}$ related questions that search users also ask, $\bm{a}$ questions and answers from crowdsourced \ssc{QA} platforms, and the snippets from the knowledge graph and answer box (if available). These evidences are arranged in sequence up to the end of the prompt. Given the models' context limit, we only keep the top $\bm{n}$ evidences (closer to the end of the prompt) after sorting them based on the corresponding date. Unless otherwise specified, we use $(\bm{o},\bm{r},\bm{a},\bm{n}, \bm{m}) = (10,2,2,5)$ for \gptthreefive, and $(\bm{o},\bm{r},\bm{a},\bm{n}, \bm{m}) = (10,3,3,10)$ for \gptfour. Additionally, we include $\bm{m} = 5$ question-answer demonstrations at the beginning of the prompt.

\subsection{Results and Discussion}
\paragraph{\bssc{FreshPrompt} significantly improves \freshqa accuracy:} 
Table~\ref{table:freshqa_strict_utd_llms} presents concrete numbers under \strict (see Appendix~\ref{appendix:utd_llms_results} for results under \ssc{Relaxed}). \seal offers large improvements over the vanilla \gptthreefive and \gptfour across the board. \gptfour + \seal achieves absolute accuracy improvements of 47\% and 31.4\% over \gptfour under \strict and \relaxed, respectively. The reduction in the absolute accuracy gap between \strict and \relaxed (from 17.8\% to 2.2\%) also suggests that \seal dramatically diminishes the presence of outdated and hallucinated answers. Unsurprisingly, the most significant improvements for both \gptthreefive and \gptfour are on the categories of fast-changing and slow-changing questions, which both concern recent knowledge. That said, questions about old knowledge also benefit from \seal. For example, \gptfour + \seal yields a +30.5\% higher accuracy than \gptfour on questions with valid premises that involve knowledge before 2022 (+9.9\% under \relaxed). Additionally, \seal produces notable gains on false-premise questions (+37.1\% and +8.1\% respective accuracy improvements under \strict and \relaxed for \gptfour). 
\begin{table}[t]
\centering
\footnotesize
\caption{Accuracy of different search engine-augmented \llms on \freshqa under \strict (no hallucination) evaluations. \hl{Models benchmarked on the same date of April 26, 2023}. We report accuracy across different categories of questions, including \textit{fast-changing} (\textit{fast}), \textit{slow-changing} (\textit{slow}), \textit{never-changing} (\textit{never}), false-premise, questions that involve knowledge before 2022 ($< 2022$) and since 2022 ($\geq 2022$), one-hop (\textit{1-hop}) and multi-hop (\textit{$m$-hop}) questions. $^{+}$ indicates a model with access to the current date. \ssc{UTD} stands for ``up-to-date''.}%
\begin{adjustbox}{max width=\textwidth}
\begin{tabular}{l c c c c c c c c c c c c c}
\toprule
\multirow{2}{*}{\textbf{Model (size)}} & \multirow{2}{*}{\textbf{knowl.}} & \multirow{2}{*}{\textbf{all}} & \multicolumn{8}{c}{\textbf{valid premise}} & \multicolumn{2}{c}{\textbf{false premise}} \\
\cmidrule(l){4-11} \cmidrule(l){12-13} 
& \textbf{cutoff} & & {all} & {fast} & {slow} & {never} & {\mbox{\scriptsize $< 2022$}} & {\mbox{\scriptsize $\geq 2022$}} & {1-hop} & {$m$-hop} & {all} & {\mbox{\scriptsize $< 2022$}}  \\
\multicolumn{13}{l}{\emph{comparison against baselines}} \\
\textcolor{mygreen}{\ssc{Google Search (N/A)}} & UTD & 39.6 & 48.9 & 32.0 & 46.4 & 68.3 & 67.4 & 37.9 & 55.6 & 32.4 & 11.3 & 9.7 \\
\\
\textcolor{mygreen}{\gptthreefive (N/A)} & 2021 & 26.0 & 26.1 & 4.0 & 15.2 & 58.7 & 61.0 & 5.1 & 28.0 & 21.3 & 25.8 & 34.4 \\
\textcolor{mygreen}{\gptthreefive + \ssc{Self-Ask} (N/A)} & UTD & 41.6 & 51.1 & 36.8 & 43.2 & 73.0 & 73.8 & 37.4 & 52.2 & 48.1 & 12.9 & 17.2 \\
\textcolor{mygreen}{\gptthreefive + \seal} & UTD & 56.0 & 62.5 & 46.4 & 60.8 & 80.2 & 71.6 & 57.0 & 68.7 & 47.2 & 36.3 & 43.0 \\
\textcolor{mygreen}{\ssc{PPLX.AI (N/A)}} & UTD & 52.2 & 57.2 & 38.4 & 53.6 & 79.4 & 73.0 & 47.7 & 63.8 & 40.7 & 37.1 & 38.7 \\
\\
\textcolor{mygreen}{\gptfour (N/A)} & 2021$^{+}$ & 28.6 & 26.9 & 12.0 & 4.0 & 64.3 & 58.2 & 8.1 & 27.2 & 25.9 & 33.9 & 41.9 \\
\textcolor{mygreen}{\gptfour + \ssc{Self-Ask} (N/A)} & UTD & 47.8 & 47.1 & 39.2 & 46.4 & 55.6 & 51.8 & 44.3 & 43.7 & 55.6 & 50.0 & 61.3 \\
\textcolor{mygreen}{\gptfour + \freshprompt} & UTD & \textbf{75.6} & \textbf{77.1} & \textbf{59.2} & \textbf{77.6} & \textbf{94.4} & \textbf{88.7} & \textbf{70.2} & \textbf{81.3} & \textbf{66.7} & \textbf{71.0} & \textbf{77.4} \\
\midrule
\multicolumn{13}{l}{\emph{sensitivity and ablation studies}} \\
\textcolor{mygreen}{\gptthreefive (N/A)} & 2021 & 26.0 & 26.1 & 4.0 & 15.2 & 58.7 & 61.0 & 5.1 & 28.0 & 21.3 & 25.8 & 34.4 \\
\textcolor{mygreen}{\gptthreefive + \seal} & UTD & 56.0 & 62.5 & 46.4 & 60.8 & 80.2 & 71.6 & 57.0 & 68.7 & 47.2 & 36.3 & 43.0 \\
\textcolor{mygreen}{w/ \ssc{premise check}} & UTD &  35.2 & 27.1 & 14.4 & 28.0 & 38.9 & 36.2 & 21.7 & 31.0 & 17.6 & 59.7 & 67.7 \\
\\
\textcolor{mygreen}{\gptfour (N/A)} & 2021$^{+}$ & 28.6 & 26.9 & 12.0 & 4.0 & 64.3 & 58.2 & 8.1 & 27.2 & 25.9 & 33.9 & 41.9 \\
\\
\textcolor{mygreen}{\gptfour w/ \ssc{snippets only \& search order}} & UTD & 74.0 & 75.5 & 56.8 & 75.2 & 94.4 & 87.9 & 68.1 & 79.9 & 64.8 & 69.4 & 77.4 \\
\textcolor{mygreen}{\gptfour w/ \ssc{snippets only \& time order}} & UTD & 
74.8 & 75.5 & 58.4 & 74.4 & 93.7 & 87.9 & 68.1 & 79.9 & 64.8 & 72.6 & \textbf{82.8} \\
\textcolor{mygreen}{\gptfour w/ \ssc{snippets only \& random order}} & UTD & 72.4 & 73.7 & 56.8 & 69.6 & 94.4 & 87.9 & 65.1 & 78.4 & 62.0 & 68.5 & 76.3 \\
\\
\textcolor{mygreen}{\gptfour + \freshprompt} & UTD & 75.6 & 77.1 & 59.2 & 77.6 & 94.4 & \textbf{88.7} & 70.2 & 81.3 & 66.7 & 71.0 & 77.4 \\
\textcolor{mygreen}{w/ \ssc{premise check}} & UTD & 75.0 & 74.2 & 56.8 & 76.0 & 89.7 & 85.1 & 67.7 & 79.5 & 61.1 & \textbf{77.4} & 79.6 \\
\textcolor{mygreen}{w/o \ssc{answer box}} & UTD & 74.2 & 74.7 & 57.6 & 74.4 & 92.1 & \textbf{88.7} & 66.4 & 79.1 & 63.9 & 72.6 & 78.5 \\
\textcolor{mygreen}{w/o \ssc{answer box} \& \ssc{relevant info}} & UTD & 72.4 & 72.9 & 54.4 & 71.2 & 92.9 & 87.2 & 64.3 & 78.0 & 60.2 & 71.0 & 78.5 \\
\textcolor{mygreen}{w/ \ssc{1 evidence}} & UTD & 61.4 & 60.9 & 40.0 & 55.2 & 87.3 & 79.4 & 49.8 & 66.8 & 46.3 & 62.9 & 75.3 \\
\textcolor{mygreen}{w/ \ssc{5 evidences}} & UTD &  70.6 & 72.1 & 56.0 & 69.6 & 90.5 & 81.6 & 66.4 & 78.0 & 57.4 & 66.1 & 73.1 \\
\textcolor{mygreen}{w/ \ssc{15 evidences}} & UTD & \textbf{77.6} & \textbf{78.5} & \textbf{60.8} & \textbf{78.4} & \textbf{96.0} & \textbf{88.7} & \textbf{72.3} & \textbf{81.7} & \textbf{70.4} & 75.0 & 80.6 \\
\textcolor{mygreen}{w/ \ssc{15 demonstrations}} & UTD & 74.6 & 75.5 & 56.8 & 76.0 & 93.7 & 87.9 & 68.1 & 79.9 & 64.8 & 71.8 & 76.3 \\
\textcolor{mygreen}{w/ \ssc{long demonstration answers}} & UTD & 73.0 & 72.6 & 55.2 & 71.2 & 91.3 & 83.7 & 66.0 & 77.6 & 60.2 & 74.2 & 81.7 \\
\bottomrule
\end{tabular}
\end{adjustbox}
\label{table:freshqa_strict_utd_llms}
\vspace{-3mm}
\end{table}

\vspace{-1em}
\paragraph{\bssc{FreshPrompt} outperforms other search-augmented methods by a large margin:} \gptfour + \seal demonstrates superior accuracy across question types, surpassing all other methods by a substantial margin. Its best variant (with 15 retrieved evidences per question) achieves impressive overall accuracies of 77.6\% and 79.0\% under \strict and \relaxed, respectively. \gptthreefive + \seal surpasses \ssc{PPLX.AI} and \ssc{Self-ask} (all performed on top of \gptthreefive) in overall accuracy by +3.8\% and +14.4\% under \strict. Under \relaxed, however, \ssc{PPLX.AI} achieves a +4.2\% higher overall accuracy than \gptthreefive + \seal, which is a large part due to its superior accuracy on false-premise questions (58.1\% vs. 41.1\%). The large accuracy gap of 14.0\% between \strict and \relaxed for \ssc{PPLX.AI} suggests that its outputs contain a large amount of hallucination. Overall, all search-engine augmented approaches (\ssc{Self-ask}, \ssc{PPLX.AI}, and \seal) provide significant gains across question types over vanilla \gptthreefive and \gptfour.
\ssc{Google Search} generally provides better results than both \gptthreefive and \gptfour, except on questions with false premises, but lags far behind \ssc{PPLX.AI} and \gptthreefive/\gptfour + \freshprompt across the board.
\vspace{-1em}
\paragraph{The premise check boosts accuracy on false-premise questions but can hurt accuracy on those with valid premises:} 
As discussed in Section~\ref{subsection:ood_llms_results}, \ssc{OpenAI}'s \llms such as \gptthreefive and \gptfour are likely tuned to handle false-premise questions, and this is also true
for \ssc{PPLX.AI}. Additionally, we empirically find that several \llms possess the ability to debunk a false-premise question if explicitly asked, e.g.. \textit{``Please check if the
question contains a valid premise before answering''}. Adding this premise check to \gptthreefive and \gptfour yields +23.4\% and +6.4\% respective accuracy improvement on false-premise questions under \strict (+22.6\% and +11.3\% under \relaxed). However, this is harmful for \gptthreefive with regard to other question types,  decreasing overall accuracy by 20.8\% and 21\% under \strict and \relaxed, respectively. This is not a problem for \gptfour, with a slight decrease of 0.6\% under \strict and a slight increase of 
and 1.2\% under \relaxed.
\vspace{-1em}
\paragraph{Having more relevant and up-to-date evidences at the end of the input context is helpful:} We also analyze how the order of the evidences in the prompt impacts \gptfour's accuracy. Our results show that using the order returned by \ssc{Google Search} (\ssc{search order}, top search results at the end of the input context) or sorting the evidences by their associated date information (\ssc{time order}, more recent results at the end) generally results in better accuracy compared to using a random order (\ssc{random order}), with up to a +2.2\% higher overall accuracy in \strict and \relaxed. Using only the text snippet for each evidence without additional information (such as source, date, etc.) as in \gptfour + \seal slightly reduces accuracy, with less than 1\% in both settings.
\vspace{-1em}
\paragraph{Additional retrieved information beyond the organic search results provides further gains:}
Incorporating additional retrieved evidences other than the \textit{organic search results}, such as the \textit{answer box} or \textit{related questions} that search users also ask, is helpful. Removing the \textit{answer box} decreases \gptfour + \seal's overall accuracy under \strict by 1.4\% (1.6\% under \relaxed). Removing both the \textit{answer box} and other relevant information (including \textit{related questions}) reduces \gptfour + \seal's overall accuracy by 3.2\% (3.0\% under \relaxed).
\vspace{-1em}
\paragraph{Increasing the number of retrieved evidences further improves \bssc{FreshPrompt}:} We explore the effect of the number of retrieved evidences for each question as well as the number of demonstrations by varying these numbers in our experiments with \gptfour. Note that our default setting for \gptfour + \seal uses 10 retrieved evidences for each question and 5 demonstrations. Our results suggest that the number of retrieved evidences for each question is the most important ingredient for achieving highest accuracy. Under \strict, increasing this number from 1 to 5, 10, and 15 leads to corresponding overall accuracy improvements of +9.2\%, +14.2\%, and +16.2\%, respectively. This suggests that \gptfour is able to efficiently handle an increasing number of retrieved evidences (including conflicting answers) and ground its responses into the most factual and up-to-date information. On the other hand, increasing the number of \emph{demonstrations} from 5 to 15 slightly hurts accuracy in both evaluation settings (1\% decrease in overall accuracy under \strict).
\vspace{-1em}
\paragraph{Verbose demonstrations improve on complex questions but also increase hallucination:} To evaluate the effect of the writing style of the answer (including the reasoning) in each demonstration, we manually rewrite these answers into a more verbose version (\ssc{long demonstration answers}). Our manual inspection reveals that using more verbose demonstration answers may be helpful when dealing with complex questions but can be more harmful as it provides room for hallucination (a decrease of 2.6\% in overall accuracy under \strict).

\section{Related Work}

\paragraph{Knowledge augmented \bssc{LLMs}:} Many prior works study semi-parametric knowledge augmentation in \llms via additional fine-tuning~\citep{KGuu20,PLewis20,SBorgeaud22,GIzacard22}, while others advocate for knowledge generation instead of retrieval~\citep{WYu23a,ZSun23}. %
\freshprompt aligns with a recent emerging trend in \ssc{QA} applications that augments \llms' prompts with knowledge retrieved from search engines for real-time alignment to current and factual information~\citep{RNakano21,ALazaridou22,JMenick22,SYao22,OPress22,OKhattab22,TSchick23,HLuo23}. Similar to our method, \citet{ALazaridou22} proposed a few-shot in-context learning approach that inserts documents from \google into the prompt. We do not compare to this method due to its expensive inference cost, as it chunks retrieved documents into evidence paragraphs and performs $k=50$ inference calls to the \llm to generate $k$ answers followed by \llm reranking. In contrast, \freshprompt only performs a single inference call to the \llm. 
\selfask~\citep{OPress22} also uses few-shot in-context learning to teach an \llm to ask itself follow-up questions before answering the initial question, although it focuses more on decomposition.

\vspace{-1em}
\paragraph{Time-sensitive \bssc{QA}:} \freshqa aligns with a growing body of work on benchmarking \llms' temporal reasoning capabilities~\citep{WChen21,MZhang21,ALiska22,JKasai22}. \citet{WChen21} created \ssc{TimeQA} by extracting evolving facts from \ssc{WikiData} along with aligned \ssc{Wikipedia} passages to synthesize 20\ssc{K} %
timestamped question-answer pairs.
\citet{MZhang21} constructed \ssc{SituatedQA} by annotating 9\ssc{K} realistic questions from existing open-domain \ssc{QA} datasets with temporal context (i.e., timestamps). 
\ssc{StreamingQA}~\citep{ALiska22} consists of both \llm-generated and human-written questions (146\ssc{K} total questions) answerable from a corpus of timestamped news articles.
Also related is the dynamic \ssc{RealtimeQA} benchmark~\citep{JKasai22}, which evaluates models weekly on a set of around 30 multiple-choice questions about new events extracted from news websites. In contrast,  \freshqa contains a fixed set of human written open-ended questions whose answers by nature can change based on new developments in the world and thus offers a complementary generative evaluation of time-sensitive \ssc{QA}.
\vspace{-1em}
\paragraph{\bssc{QA} over questionable or counterfactual premises:} Recent work has also introduced \ssc{QA} benchmarks with questionable premises~\citep{XYu23,NKim23} or counterfactual premises~\citep{WYu23b}. \ssc{Crepe}\citep{XYu23} consists of 8400 Reddit questions (of which 25\% questions contain false premises annotated by human workers) split into train/dev/test sets. \citet{NKim23} constructed \ssc{(QA)$^2$}, an evaluation set of 602 questions based on frequent search engine queries, which are annotated by expert annotators and crowdworkers, and evenly divided between those with and without questionable premises.
Consistent with these efforts, we find that current \llms struggle with handling false premise questions; additionally, several \ssc{LLMs} are able to debunk a false-premise question if explicitly asked to check for the premise's validity. Similar to above, these benchmarks are complementary and combining them is a promising direction for future work.

\vspace{-0.2em}
\section{Limitations and Future Work}
One obvious challenge with \freshqa is the need for regular answer updating by the maintainers; in the interim period between updates, the answers to some questions might become stale. This could be addressed by support from the open-source community (e.g., updates via \ssc{GitHub} pull requests).
On the method side, \freshprompt\ interfaces with \google, and it is unclear how it performs with other search engines for which some types of context (e.g., answer boxes) are not available. Additionally, we only perform one search query per question, and thus our method could be further improved via question decomposition and multiple search queries~\citep{OKhattab22}. Since \freshqa consists of relatively simple English language questions, it is also unclear how well \freshprompt performs in the context of multilingual/cross-lingual \ssc{QA} and long-form \ssc{QA}~\citep{AFan19}. Finally, \freshprompt\ relies on in-context learning and thus  may underperform approaches that fine-tune the base \llm\ on new knowledge.

\vspace{-0.22em}
\section{Conclusion}
Our work offers a fine-grained and exhaustive evaluation of the capabilities of modern \llms to adapt to ever-changing world knowledge with and without search engine augmentation. In the process, we develop a new dataset---\freshqa---of 600 questions that test a broad range of reasoning abilities, from the incorporation of fast-changing knowledge to identification of questions with false premises. Our two-mode evaluation also provides a way to measure both correctness and hallucination. Additionally, we propose a simple few-shot in-context learning algorithm called \freshprompt\ that incorporates relevant evidences retrieved from \google into the prompt of an \llm. \freshprompt\ significantly improves performance over competing search engine-augmented approaches on \freshqa, and an ablation reveals that factors such as the number of incorporated evidences and their order impact the correctness of \llm-generated answers. We release \freshqa\ and commit to updating its answers regularly to facilitate future research.
\newpage
\clearpage
\section{Acknowledgements}
We thank Colin Raffel, Hamed Zamani, and Subhransu Maji for helpful discussion and feedback. We would also like to thank Chengrun Yang, Xinyun Chen for their insightful comments on this manuscript. Finally, we are grateful to the following people for their contributions to creating our \freshqa dataset: Marzena Karpinska, Dustin Tran, Daniel Cer, Sam Fullerton, Elizabeth Clark, Nishant Raj, Xiaoyu Song, Yapei Chang, Yixiao Song, Nader Akoury, Ankita Gupta, Bill Ray, Chau Pham, Wenlong Zhao, Maximilian Mozes, Simeng Sun, Ronan Salz, Kalpesh Krishna, Katherine Thai, Kanishka Misra, Salaheddin Alzu'bi, Erica Cai, Thibault Sellam, Jiao Sun, Dhruv Agarwal, Tessa Masis, Andrew Drozdov, Brian Lester, George Wei, Naveen Jafer Nizar, Shufan Wang, Youngwoo Kim, and Shib Sankar Dasgupta. This project was partially supported by award \ssc{IIS}-2046248 from the National Science Foundation (\ssc{NSF}), as well as \ssc{NSF}'s \ssc{CloudBank} program.
\bibliography{iclr2024_conference}
\bibliographystyle{iclr2024_conference}
\clearpage
\newpage
\appendix
\section*{Appendix}
\label{section:appendices}
\section{Evaluation protocol}
\label{appendix:evaluation_protocol}
Figure~\ref{figure:freshqa_eval} shows specific examples of each evaluation criteria.
\section{Inter-rater agreement and automatic evaluation}
\label{appendix:inter_rater_agreement}
Two authors independently evaluated a randomly sampled subset of 100 answers across models (including 50 questions with valid premises and 50 questions with false premises) in both modes \relaxed and \strict.

\begin{figure}[h]
\centering
\includegraphics[width=\textwidth]{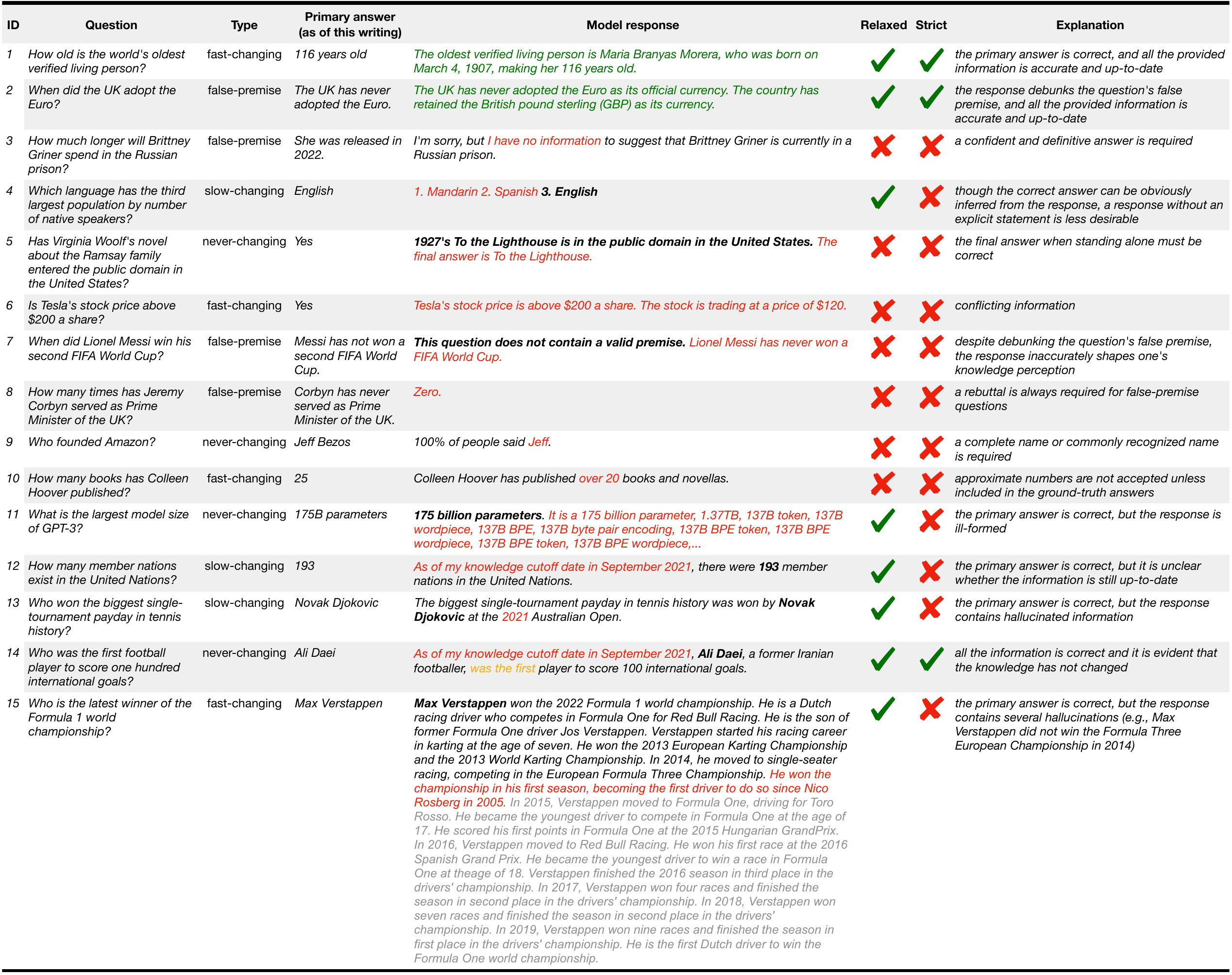}
\caption{\freshqa sample evaluation. To get credit in both evaluation modes \relaxed and \strict, all the information in the answer must be accurate and up-to-date (examples 1 and 2). In both modes, we credit a model's response only if it provides a confident and definitive answer (example 3), or the correct answer can be obviously inferred from the response (provided all other requirements are satisfied, see example 4). The primary or final answer when standing alone must be accurate (example 5). Any additional information that is provided must not contradict the primary answer (example 6) or reshape one's perception of it (example 7). For false-premise questions, the model must point out the presence of a false premise to receive credit (example 8). For answers that involve names of entities (e.g., people), complete names or commonly recognized names are expected (example 9). Regarding numerical answers, approximate numbers are generally not accepted unless explicitly included in the ground-truth answers (example 10). Under \relaxed, we accept ill-formed responses (including those in a non-English language), as well as hallucinated or outdated information that does not significantly impact the primary answer; under \strict, however, a response that contains any hallucination, no matter how minor, will not receive credit (examples 11, 12, and 13). Furthermore, we accept a response in \strict\ when the model indicates that the information might be outdated (e.g., {\textit{``As of my knowledge cutoff date in September 2021''}}) \emph{only} if it is evident that the knowledge has not changed (example 14).}
\label{figure:freshqa_eval}
\vspace{-2mm}
\end{figure}

\begin{figure}[t!]
\centering
\includegraphics[width=\textwidth]{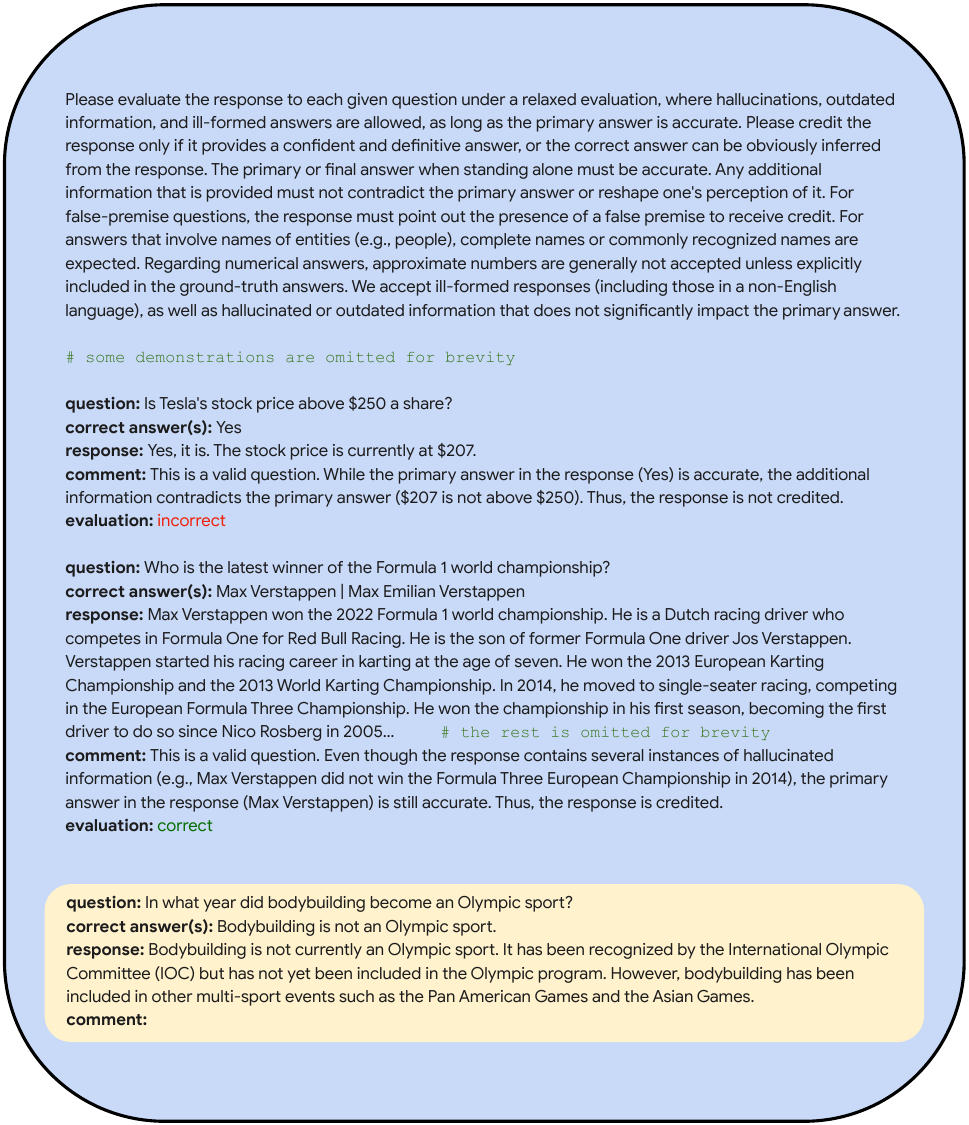}
\caption{\fresheval's prompt for \relaxed evaluation.}
\label{figure:fresheval_relaxed_prompt}
\vspace{-2mm}
\end{figure}

\begin{figure}[t!]
\centering
\includegraphics[width=\textwidth]{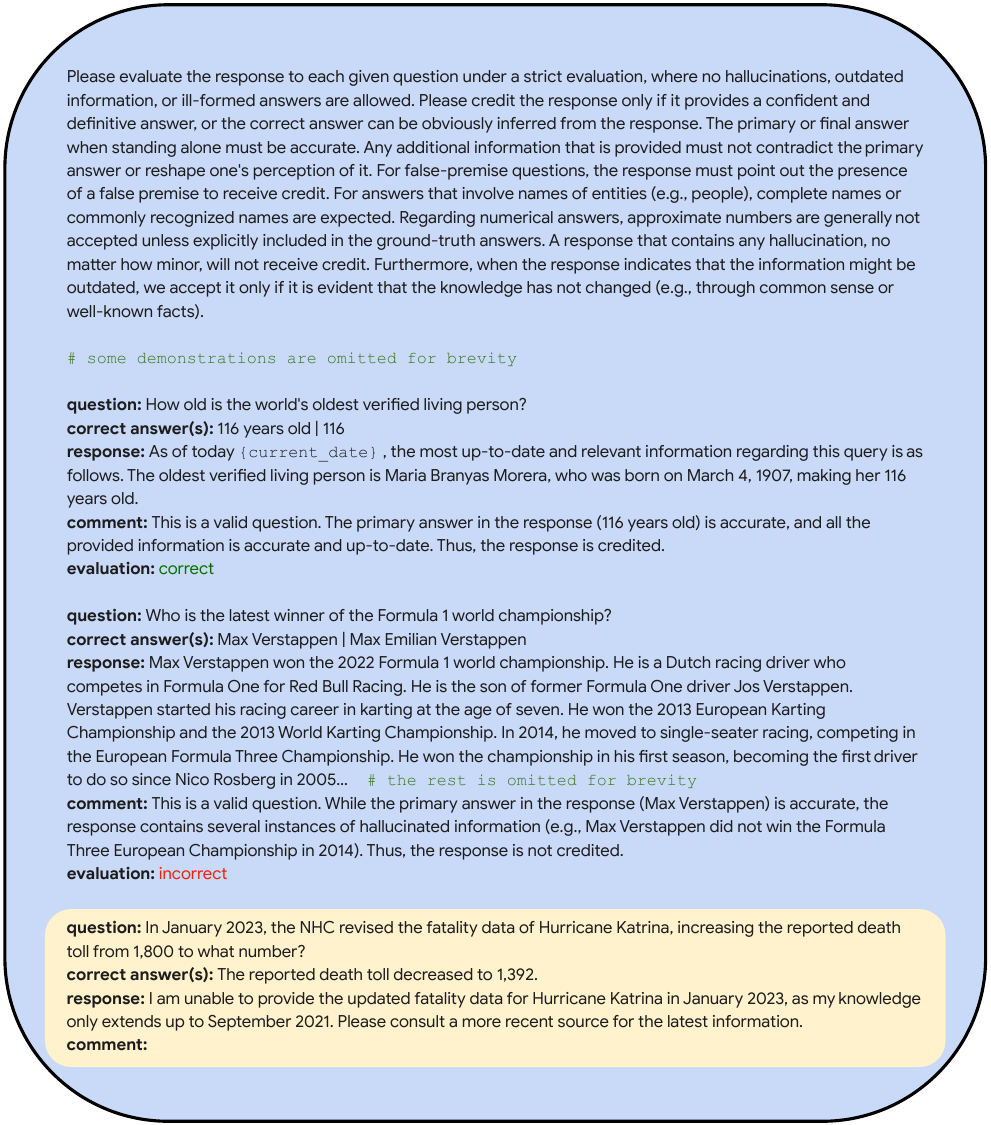}
\caption{\fresheval's prompt for \strict evaluation.}
\label{figure:fresheval_strict_prompt}
\vspace{-2mm}
\end{figure}

\begin{figure}[t!]
\centering
\includegraphics[width=\textwidth]{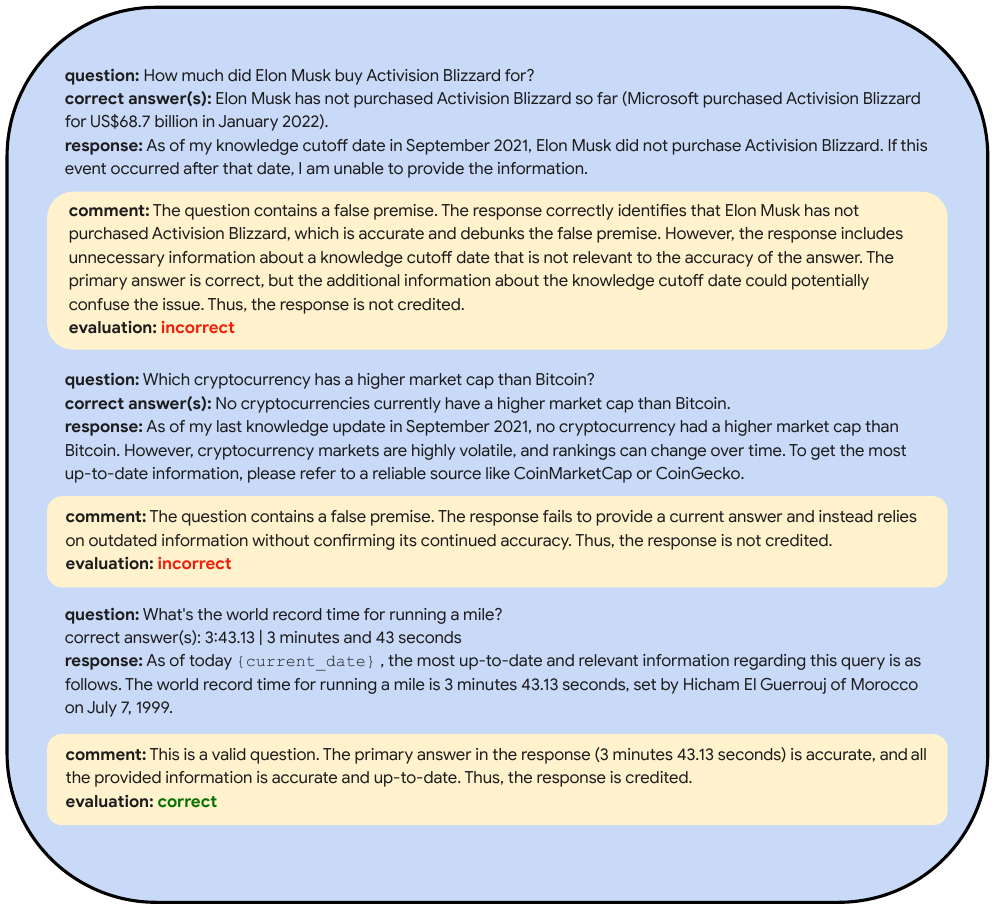}
\caption{\fresheval's sample output for \strict evaluation.}
\label{figure:fresheval_sample_output}
\vspace{-2mm}
\end{figure}

\textcolor{black}{To facilitate future evaluations, we also develop \fresheval, a simple automatic metric that uses few-shot in-context learning to teach an \llm to judge model responses. In each evaluation, the model is conditioned on a given question, a list of valid answers for the question, and a model response, and is then expected to generate a comment on the correctness of the response, followed by a final judgement. At the beginning of each input prompt, we also provide an instruction of the evaluation task, and sample comments and evaluations of the examples in Figure~\ref{figure:freshqa_eval} as demonstrations.\footnote{\textcolor{black}{In our experiments, we found that using separate prompts for \fssc{Relaxed} and \fssc{Strict} evaluations resulted in better performance compared to using a single, combined prompt for both evaluation modes. We also found that additionally incorporating retrieved evidences for the question into the prompt did not improve inter-rater agreement between \fssc{FreshEval} and human raters.}} See Figure~\ref{figure:fresheval_relaxed_prompt} and Figure~\ref{figure:fresheval_strict_prompt} for \fresheval's prompts for \relaxed and \strict evaluations, and Figure~\ref{figure:fresheval_sample_output} for \fresheval's sample output for \strict evaluation.}

\textcolor{black}{Table~\ref{table:inter_rater_agreement} reports the inter-rater agreement between the two human raters, and between \fresheval and each human rater, in terms of exact accuracy. The two human raters had an agreement of 99\% for \relaxed\ and 96\% for \strict, while \fresheval achieved an average agreement of 96.5\% with human evaluations for \relaxed\ and 96\% for \strict. Overall, the high accuracies demonstrate that our evaluation protocol is reproducible and reliable, and \fresheval can be used in place of human evaluation on \freshqa.}
\section{Additional experiment setup details for %
Section~\ref{section:ood_llms}}
\label{appendix:ood_llms_details}
To increase reproducibility, we select the most likely token at every decoding timestep (i.e., with a temperature of 0) and generate a maximum number of 256 tokens for all models. Note that the \ssc{API} for some models is non-deterministic by default, even with a temperature of 0. For non-chat models that were not pre-trained with a \ssc{QA} task, we feed them a text prompt of the format: ``Q: \texttt{<question>}\textbackslash nA: '' (``\texttt{\textbackslash} n'' is the new line character). 

For \ssc{OpenAI} models, we use the \texttt{2023-03-15-preview} \ssc{API} in \ssc{Azure OpenAI Service}. We use the model names \texttt{text-davinci-003}, \texttt{code-davinci-002}, \texttt{gpt-3.5-turbo}, and \texttt{gpt-4} for \ssc{GPT-3.5}, \ssc{Codex}, \ssc{ChatGPT}, and \ssc{GPT-4}, respectively.

\section{Additional experiment results for Section~\ref{section:ood_llms}}
\label{appendix:ood_llms_results}
Table~\ref{table:freshqa_strict_ood_llms} and Table~\ref{table:freshqa_relaxed_ood_llms} show the accuracy of different \llms on \freshqa under \strict (no hallucination) and \relaxed evaluations, respectively.

\section{\textsc{ChatGPT/GPT-4}'s awareness of recent knowledge}
\label{appendix:chatgpt_gpt4_recent_knowledge}
\begin{figure}[t!]
\centering
\includegraphics[width=0.8\textwidth]{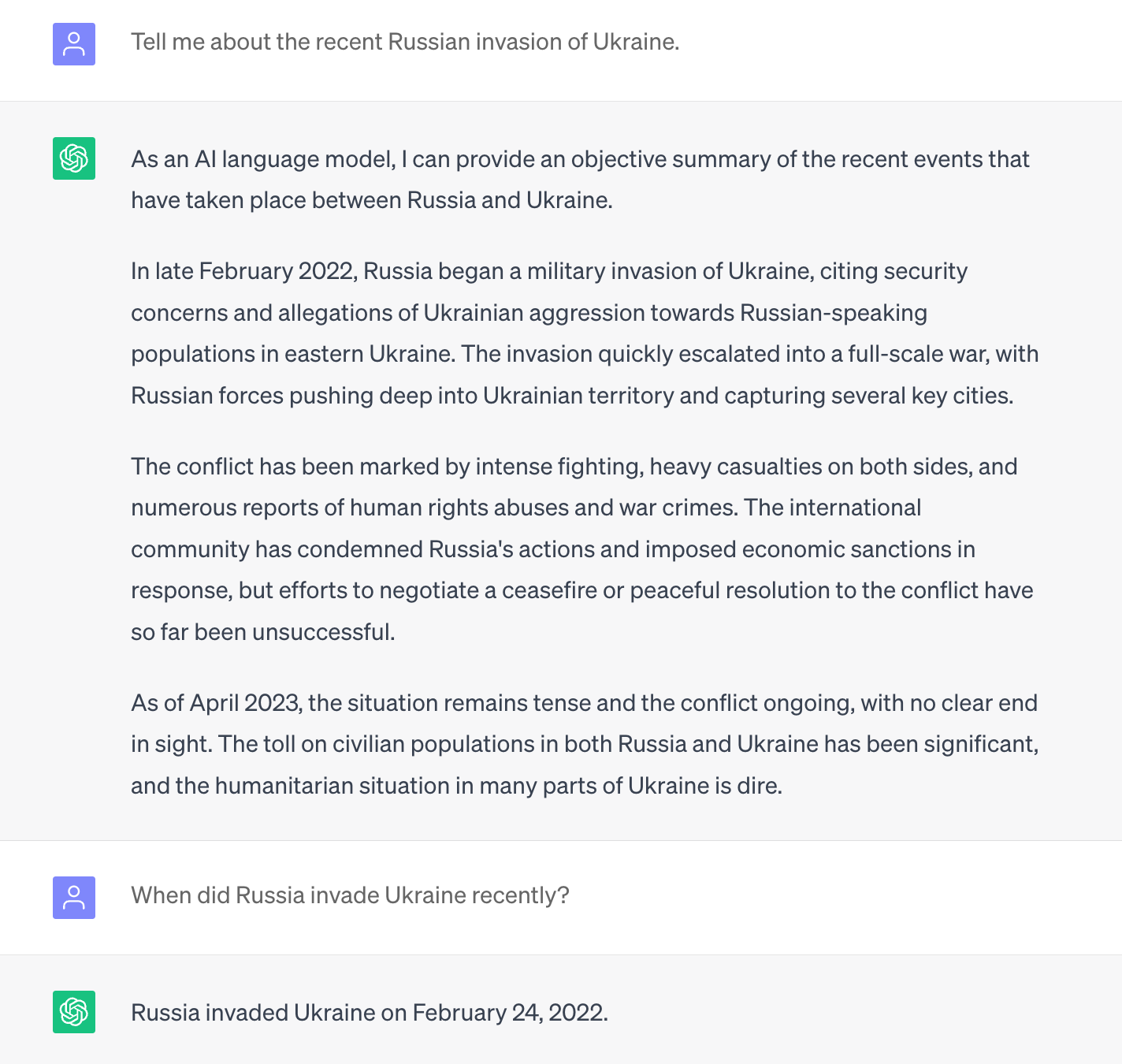}
\caption{Despite its knowledge cutoff date in September 2021, \ssc{ChatGPT} is aware of the recent Russian invasion of Ukraine on February 24, 2022. \hl{Questions asked on April 9, 2023}.}
\label{figure:chatgpt_recent_knowledge}
\vspace{-2mm}
\end{figure}

Although \ssc{ChatGPT} and \ssc{GPT-4} were originally trained in 2021, our manual evaluation suggests that they have been exposed to data containing information beyond their knowledge cutoff date in September, 2021.  Figure~\ref{figure:chatgpt_recent_knowledge} indicates that \ssc{ChatGPT} is aware of the recent Russian invasion of Ukraine on February 24, 2022.

\section{\textsc{Google Search} results}
\begin{figure}[t!]
\centering
\includegraphics[width=\textwidth]{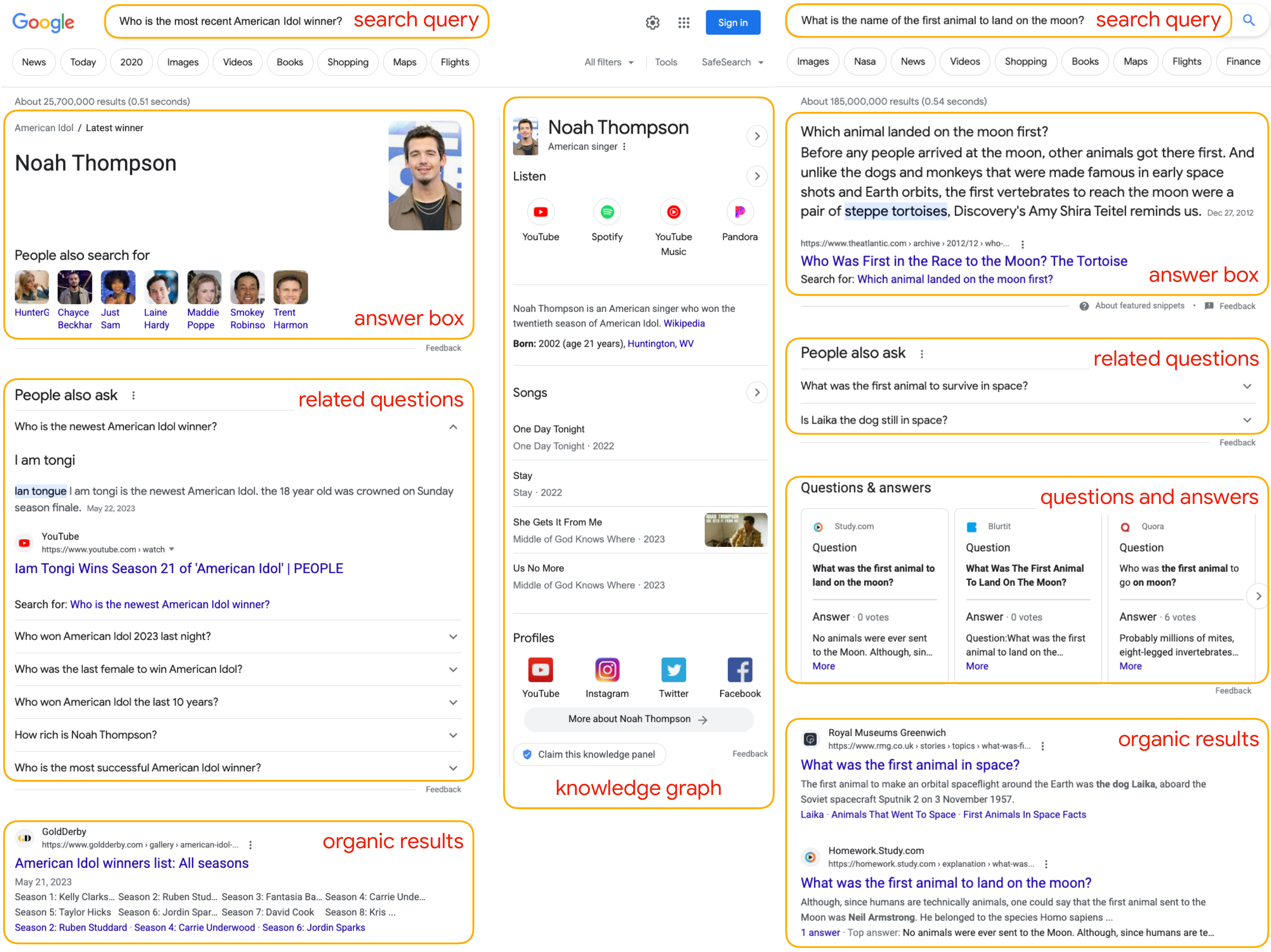}
\caption{\google produces different types of search results for given a query, including the \textit{answer box}, \textit{organic results}, and other useful information, such as the \textit{knowledge graph}, \textit{questions and answers} from crowdsourced \ssc{QA} platforms, and \textit{related questions} that search users also ask. Each of these results contains an associated \textit{text snippet} along with other information, such as \textit{source webpage}, \textit{date}, \textit{title}, and \textit{highlighted words}.}
\label{figure:search_results}
\vspace{-2mm}
\end{figure}

Figure~\ref{figure:search_results} shows different types of search results from \google for given a query.
\label{appendix:search_results}

\section{A realistic prompt for FreshPrompt}
\label{appendix:freshprompt_realistic_prompt}
\begin{figure}[t!]
\centering
\includegraphics[width=\textwidth]{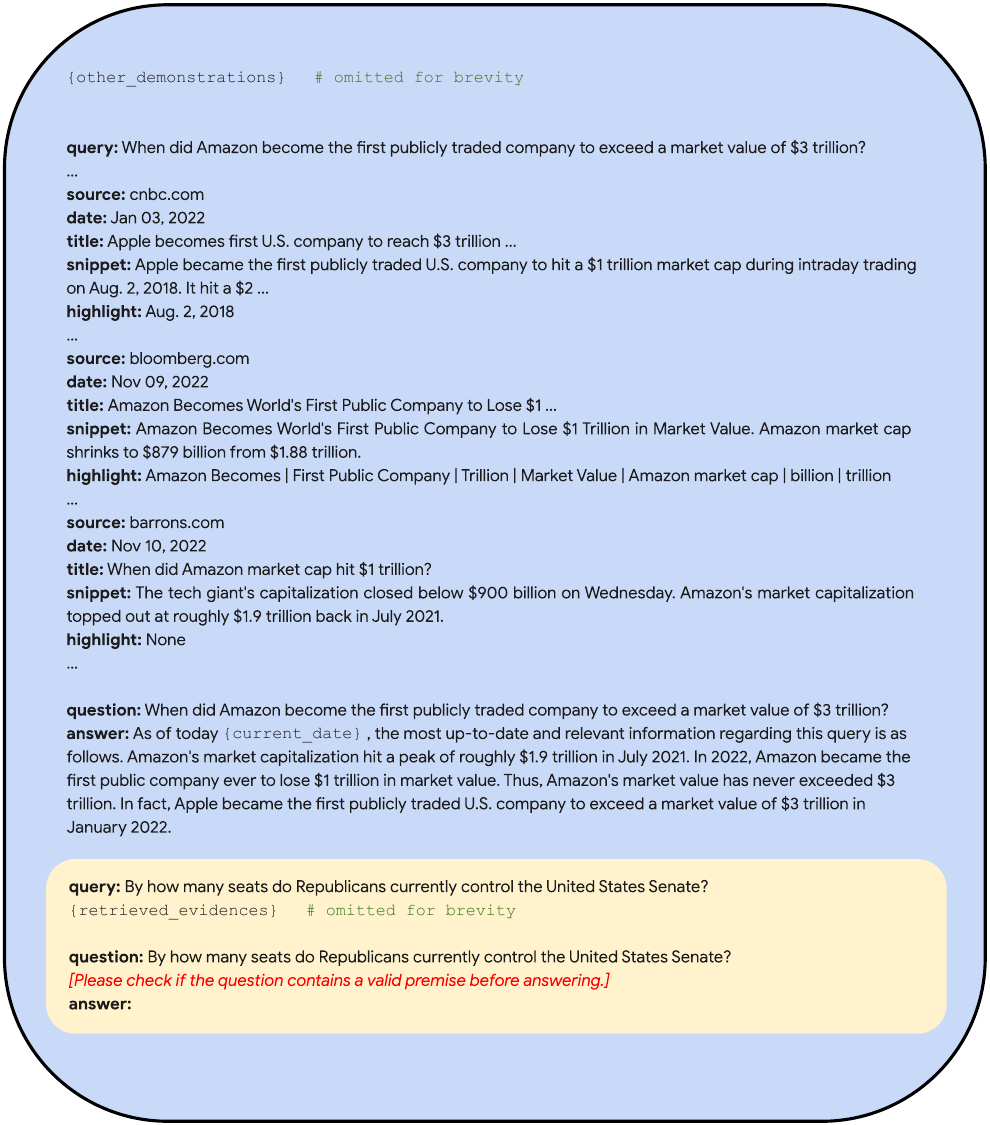}
\caption{A realistic prompt for \freshprompt. We cast all retrieved evidences into a unified format with useful information, including source webpage, date, title, text snippet, and highlighted words. Few-shot demonstrations are provided at the beginning of the prompt. Each demonstration shows the model an example question and a list of retrieved evidences for the question, followed by some reasoning over the evidences to figure out the most relevant and up-to-date answer.}
\label{figure:freshprompt_realistic_prompt}
\vspace{-2mm}
\end{figure}

Figure~\ref{figure:freshprompt_realistic_prompt} displays a realistic prompt for \freshprompt.

\section{Additional experiment results for Section~\ref{section:utd_llms}}
\label{appendix:utd_llms_results}
Table~\ref{table:freshqa_relaxed_utd_llms} presents the accuracy of different search engine-augmented \llms on \freshqa under \relaxed.

\clearpage
\newpage
\begin{table}[ht!]
\centering
\footnotesize
\caption{Inter-rater agreement between two authors (\ssc{Rater 1} and \ssc{Rater 2}), and between \fresheval and each human rater, in terms of exact accuracy across 100 \relaxed judgements, 100 \strict judgements, and all \ssc{All} 200 judgements. In each of these three categories, in addition to the overall accuracy (\textbf{overall}), we report accuracy across questions with valid premises (\textbf{vp}) and questions with false premises (\textbf{fp}). The high accuracies demonstrate that our evaluation protocol is reproducible and reliable, and \fresheval can be used in place of human evaluation on \freshqa.}%
\begin{adjustbox}{max width=\textwidth}
\begin{tabular}{l c c c c c c c c c }
\toprule
 & \multicolumn{3}{c}{\relaxed} & \multicolumn{3}{c}{\strict} & \multicolumn{3}{c}{\ssc{All}} \\
\cmidrule(l){2-4} \cmidrule(l){5-7} \cmidrule(l){8-10} 
& \textbf{overall}  & \textbf{vp} & \textbf{fp} & \textbf{overall} & \textbf{vp} & \textbf{fp} & \textbf{overall} & \textbf{vp} & \textbf{fp}\\[2mm]
\ssc{Rater 1} vs. \ssc{Rater 2} & \textbf{99.0} & \textbf{98.0} & \textbf{100} & 96.0 & \textbf{100.0} & 92.0 & \textbf{97.5} & \textbf{99.0} & \textbf{96.0}\\
\fresheval vs. \ssc{Rater 1} & 97.0 & \textbf{98.0} & 96.0 & \textbf{97.0} & \textbf{100.0} & \textbf{94.0} & 97.0 & \textbf{99.0} & 95.0\\
\fresheval vs. \ssc{Rater 2} & 96.0 & 96.0 & 96.0 & 95.0 & \textbf{100.0} & 90.0 & 95.5 & 98.0 & 93.0\\
\bottomrule
\end{tabular}
\end{adjustbox}
\label{table:inter_rater_agreement}
\vspace{-3mm}
\end{table}

\clearpage
\newpage
\clearpage
\newpage
\begin{table}[t!]
\centering
\caption{Accuracy of different \llms on \freshqa under \strict (no hallucination) evaluations. \hl{Models benchmarked on the same date of April 26, 2023}. We report accuracy across different categories of questions, including \textit{fast-changing} (\textit{fast}), \textit{slow-changing} (\textit{slow}), \textit{never-changing} (\textit{never}), false-premise, questions that involve knowledge before 2022 ($< 2022$) and since 2022 ($\geq 2022$), one-hop (\textit{1-hop}) and multi-hop (\textit{$m$-hop}) questions. $^{+}$ indicates a model with access to the current date.}%
\begin{adjustbox}{max width=\textwidth}
\begin{tabular}{l c c c c c c c c c c c c c}
\toprule
\multirow{2}{*}{\textbf{Model (size)}} & \multirow{2}{*}{\textbf{knowl.}} & \multirow{2}{*}{\textbf{all}} & \multicolumn{8}{c}{\textbf{valid premise}} & \multicolumn{2}{c}{\textbf{false premise}} \\
\cmidrule(l){4-11} \cmidrule(l){12-13} 
& \textbf{cutoff} & & {all} & {fast} & {slow} & {never} & {\mbox{\scriptsize $< 2022$}} & {\mbox{\scriptsize $\geq 2022$}} & {1-hop} & {$m$-hop} & {all} & {\mbox{\scriptsize $< 2022$}}  \\
\multicolumn{13}{l}{\emph{without access to a search engine}} \\
\textcolor{mygreen}{\ssc{OpenAI Codex (N/A)}} & 2021 & 25.0 & \textbf{31.4} & 5.6 & \textbf{28.0} & 60.3 & \textbf{64.5} & 11.5 & \textbf{34.7} & 23.1 & 5.6 & 7.5 \\
\textcolor{mygreen}{\ssc{GPT 3.5 (N/A)}} & 2021 & 26.0 & 26.1 & 4.0 & 15.2 & 58.7 & 61.0 & 5.1 & 28.0 & 21.3 & 25.8 & 34.4 \\
\textcolor{mygreen}{\ssc{ChatGPT (N/A)}} & 2021$^{+}$ & \textbf{32.0} & 28.5 & 7.2 & 16.0 & 61.9 & 63.1 & 7.7 & 29.9 & 25.0 & \textbf{42.7} & \textbf{52.7} \\
\textcolor{mygreen}{\ssc{GPT 4 (N/A)}} & 2021$^{+}$ & 28.6 & 26.9 & \textbf{12.0} & 4.0 & \textbf{64.3} & 58.2 & 8.1 & 27.2 & 25.9 & 33.9 & 41.9  \\
\cmidrule(l){1-1}
\textcolor{mygreen}{\ssc{FLAN-PaLM (540B)}} & 2022 & 23.4 & 30.3 & 10.4 & 24.8 & 55.6 & 60.3 & \textbf{12.3} & 32.5 & 25.0 & 2.4 & 3.2 \\
\cmidrule(l){1-1}
\textcolor{mygreen}{\ssc{PaLM (540B)}} & 2021 & 7.2 & 9.3 & 0.8 & 11.2 & 15.9 & 20.6 & 2.6 & 9.3 & 9.3 & 0.8 & 1.1 \\
\textcolor{mygreen}{w/ \ssc{Few-shot}} & & 20.0 & 26.3 & 5.6 & 19.2 & 54.0 & 56.7 & 8.1 & 25.7 & \textbf{27.8} & 0.8 & 1.1 \\
\textcolor{mygreen}{w/ \ssc{CoT}} & & 15.4 & 19.1 & 0.8 & 9.6 & 46.8 & 47.5 & 2.1 & 20.5 & 15.7 & 4.0 & 5.4 \\
\cmidrule(l){1-1}
\textcolor{mygreen}{\ssc{PaLMChilla (62B)}} & 2022 & 12.2 & 16.0 & 2.4 & 15.2 & 30.2 & 35.5 & 4.3 & 17.2 & 13.0 & 0.8 & 1.1 \\
\cmidrule(l){1-1}
\textcolor{mygreen}{\ssc{PaLM (62B)}} & 2021 & 6.2 & 8.2 & 1.6 & 8.8 & 14.3 & 16.3 & 3.4 & 7.8 & 9.3 & 0.0 & 0.0 \\
\textcolor{mygreen}{w/ \ssc{Few-shot}} & & 12.8 & 16.8 & 3.2 & 15.2 & 31.7 & 35.5 & 5.5 & 17.9 & 13.9 & 0.8 & 1.1 \\
\textcolor{mygreen}{w/ \ssc{CoT}} & & 7.0 & 9.0 & 0.8 & 6.4 & 19.8 & 21.3 & 1.7 & 10.1 & 6.5 & 0.8 & 1.1 \\
\cmidrule(l){1-1}
\textcolor{mygreen}{\ssc{PaLM (8B)}} & 2021 & 5.6 & 7.5 & 0.8 & 5.6 & 16.0 & 16.2 & 2.1 & 8.6 & 4.6 & 0.0 & 0.0 \\
\textcolor{mygreen}{w/ \ssc{Few-shot}} & & 8.4 & 11.2 & 0.8 & 9.6 & 23.0 & 24.8 & 3.0 & 14.2 & 3.7 & 0.0 & 0.0 \\
\textcolor{mygreen}{w/ \ssc{CoT}} & & 7.8 & 10.4 & 0.0 & 6.4 & 24.6 & 24.8 & 1.7 & 11.2 & 8.3 & 0.0 & 0.0 \\
\cmidrule(l){1-1}
\textcolor{mygreen}{\ssc{FLAN-T5 XXL (11B)}} & 2022 & 6.6 & 8.8 & 3.2 & 10.4 & 12.7 & 13.5 & 6.0 & 10.1 & 5.6 & 0.0 & 0.0 \\
\cmidrule(l){1-1}
\textcolor{mygreen}{\ssc{T5 XXL (11B)}} & 2019 & 7.0 & 8.8 & 2.4 & 4.8 & 19.0 & 16.3 & 4.3 & 10.4 & 4.6 & 1.6 & 2.2 \\
\textcolor{mygreen}{w/ \ssc{Few-shot}} & & 8.4 & 11.2 & 5.6 & 11.2 & 16.7 & 17.7 & 7.2 & 13.4 & 5.6 & 0.0 & 0.0 \\
\textcolor{mygreen}{w/ \ssc{CoT}} & & 6.2 & 8.2 & 2.4 & 6.4 & 15.9 & 15.6 & 3.8 & 8.6 & 7.4 & 0.0 & 0.0 \\
\cmidrule(l){1-1}
\textcolor{mygreen}{\ssc{T5 XL (3B)}} & 2019 & 4.4 & 5.9 & 2.4 & 4.8 & 10.3 & 10.6 & 3.0 & 7.5 & 1.9 & 0.0 & 0.0 \\
\textcolor{mygreen}{w/ \ssc{Few-shot}} & & 6.0 & 8.0 & 4.0 & 8.8 & 11.1 & 13.5 & 4.7 & 8.2 & 7.4 & 0.0 & 0.0 \\
\textcolor{mygreen}{w/ \ssc{CoT}} & & 2.8 & 3.7 & 2.4 & 1.6 & 7.1 & 7.8 & 1.3 & 4.1 & 2.8 & 0.0 & 0.0 \\
\cmidrule(l){1-1}
\textcolor{mygreen}{\ssc{T5 Large (770M)}} & 2019 & 2.6 & 3.5 & 0.8 & 4.0 & 5.6 & 5.7 & 2.1 & 3.7 & 2.8 & 0.0 & 0.0 \\
\textcolor{mygreen}{w/ \ssc{Few-shot}} & & 0.8 & 1.1 & 0.0 & 0.0 & 3.2 & 2.8 & 0.0 & 1.1 & 0.9 & 0.0 & 0.0 \\
\textcolor{mygreen}{w/ \ssc{CoT}} & & 0.8 & 1.1 & 0.8 & 0.0 & 2.4 & 2.1 & 0.4 & 1.1 & 0.9 & 0.0 & 0.0 \\
\bottomrule
\end{tabular}
\end{adjustbox}
\label{table:freshqa_strict_ood_llms}
\vspace{-3mm}
\end{table}

\begin{table}[t]
\centering
\caption{Accuracy of different \llms on \freshqa under \relaxed evaluations. \hl{Models benchmarked on the same date of April 26, 2023}. We report accuracy across different categories of questions, including \textit{fast-changing} (\textit{fast}), \textit{slow-changing} (\textit{slow}), \textit{never-changing} (\textit{never}), false-premise, questions that involve knowledge before 2022 ($< 2022$) and since 2022 ($\geq 2022$), one-hop (\textit{1-hop}) and multi-hop (\textit{$m$-hop}) questions. $^{+}$ indicates a model with access to the current date.}%
\begin{adjustbox}{max width=\textwidth}
\begin{tabular}{l c c c c c c c c c c c c c}
\toprule
\multirow{2}{*}{\textbf{Model (size)}} & \multirow{2}{*}{\textbf{knowl.}} & \multirow{2}{*}{\textbf{all}} & \multicolumn{8}{c}{\textbf{valid premise}} & \multicolumn{2}{c}{\textbf{false premise}} \\
\cmidrule(l){4-11} \cmidrule(l){12-13} 
& \textbf{cutoff} & & {all} & {fast} & {slow} & {never} & {\mbox{\scriptsize $< 2022$}} & {\mbox{\scriptsize $\geq 2022$}} & {1-hop} & {$m$-hop} & {all} & {\mbox{\scriptsize $< 2022$}}  \\
\multicolumn{13}{l}{\emph{without access to a search engine}} \\
\textcolor{mygreen}{\ssc{OpenAI Codex (N/A)}} & 2021 & 25.6 & 32.2 & 6.4 & 29.6 & 60.3 & 66.0 & 11.9 & 35.4 & 24.1 & 5.6 & 7.5 \\
\textcolor{mygreen}{\ssc{GPT 3.5 (N/A)}} & 2021 & 32.4 & 32.4 & 8.0 & 28.0 & 61.1 & 68.1 & 11.1 & 34.7 & 26.9 & 32.3 & 43.0 \\
\textcolor{mygreen}{\ssc{ChatGPT (N/A)}} & 2021$^{+}$ & 41.4 & 36.7 & 10.4 & 32.8 & 66.7 & 76.6 & 12.8 & 36.2 & 38.0 & 55.6 & 66.7 \\
\textcolor{mygreen}{\ssc{GPT 4 (N/A)}} & 2021$^{+}$ & \textbf{46.4} & \textbf{39.6} & \textbf{14.4} & \textbf{35.2} & \textbf{69.0} & \textbf{80.9} & \textbf{14.9} & \textbf{39.2} & \textbf{40.7} & \textbf{66.9} & \textbf{83.9} \\
\cmidrule(l){1-1}
\textcolor{mygreen}{\ssc{FLAN-PaLM (540B)}} & 2022 & 23.6 & 30.3 & 10.4 & 24.8 & 55.6 & 60.3 & 12.3 & 32.5 & 25.0 & 3.2 & 4.3 \\
\cmidrule(l){1-1}
\textcolor{mygreen}{\ssc{PaLM (540B)}} & 2021 & 12.2 & 16.0 & 2.4 & 14.4 & 31.0 & 34.8 & 4.7 & 16.4 & 14.8 & 0.8 & 1.1 \\
\textcolor{mygreen}{w/ \ssc{Few-shot}} & & 20.2 & 26.3 & 5.6 & 19.2 & 54.0 & 56.7 & 8.1 & 25.7 & 27.8 & 1.6 & 2.2 \\
\textcolor{mygreen}{w/ \ssc{CoT}} & & 22.8 & 28.2 & 4.0 & 20.0 & 60.3 & 64.5 & 6.4 & 28.4 & 27.8 & 6.5 & 8.6 \\
\cmidrule(l){1-1}
\textcolor{mygreen}{\ssc{PaLMChilla (62B)}} & 2022 & 15.0 & 19.4 & 2.4 & 19.2 & 36.5 & 43.3 & 5.1 & 20.1 & 17.6 & 1.6 & 2.2 \\
\cmidrule(l){1-1}
\textcolor{mygreen}{\ssc{PaLM (62B)}} & 2021 & 8.6 & 11.2 & 2.4 & 11.2 & 19.8 & 22.0 & 4.7 & 11.6 & 10.2 & 0.8 & 1.1 \\
\textcolor{mygreen}{w/ \ssc{Few-shot}} & & 14.2 & 18.4 & 4.0 & 15.2 & 35.7 & 39.0 & 6.0 & 18.7 & 17.6 & 1.6 & 2.2 \\
\textcolor{mygreen}{w/ \ssc{CoT}} & & 12.8 & 16.2 & 2.4 & 15.2 & 31.0 & 34.8 & 5.1 & 17.5 & 13.0 & 2.4 & 3.2 \\
\cmidrule(l){1-1}
\textcolor{mygreen}{\ssc{PaLM (8B)}} & 2021 & 8.8 & 11.2 & 0.8 & 11.2 & 21.6 & 21.1 & 5.2 & 13.1 & 6.5 & 1.6 & 2.1 \\
\textcolor{mygreen}{w/ \ssc{Few-shot}} & & 9.2 & 12.2 & 0.8 & 10.4 & 25.4 & 27.0 & 3.4 & 15.3 & 4.6 & 0.0 & 0.0 \\
\textcolor{mygreen}{w/ \ssc{CoT}} & & 11.4 & 15.2 & 2.4 & 11.2 & 31.7 & 32.6 & 4.7 & 16.8 & 11.1 & 0.0 & 0.0 \\
\cmidrule(l){1-1}
\textcolor{mygreen}{\ssc{FLAN-T5 XXL (11B)}} & 2022 & 7.2 & 9.6 & 3.2 & 12.0 & 13.5 & 14.2 & 6.8 & 10.8 & 6.5 & 0.0 & 0.0 \\
\cmidrule(l){1-1}
\textcolor{mygreen}{\ssc{T5 XXL (11B)}} & 2019 & 10.8 & 13.8 & 3.2 & 12.8 & 25.4 & 22.7 & 8.5 & 16.0 & 8.3 & 1.6 & 2.2 \\
\textcolor{mygreen}{w/ \ssc{Few-shot}} & & 9.0 & 12.0 & 5.6 & 11.2 & 19.0 & 19.1 & 7.7 & 14.6 & 5.6 & 0.0 & 0.0 \\
\textcolor{mygreen}{w/ \ssc{CoT}} & & 13.0 & 17.3 & 4.0 & 17.6 & 30.2 & 31.2 & 8.9 & 19.0 & 13.0 & 0.0 & 0.0 \\
\cmidrule(l){1-1}
\textcolor{mygreen}{\ssc{T5 XL (3B)}} & 2019 & 5.8 & 7.7 & 4.0 & 5.6 & 13.5 & 13.5 & 4.3 & 9.0 & 4.6 & 0.0 & 0.0 \\
\textcolor{mygreen}{w/ \ssc{Few-shot}} & & 6.0 & 8.0 & 4.0 & 8.8 & 11.1 & 13.5 & 4.7 & 8.2 & 7.4 & 0.0 & 0.0 \\
\textcolor{mygreen}{w/ \ssc{CoT}} & & 5.2 & 6.9 & 3.2 & 4.0 & 13.5 & 14.2 & 2.6 & 8.6 & 2.8 & 0.0 & 0.0 \\
\cmidrule(l){1-1}
\textcolor{mygreen}{\ssc{T5 Large (770M)}} & 2019 & 4.4 & 5.3 & 2.4 & 4.8 & 8.7 & 7.1 & 4.3 & 5.6 & 4.6 & 1.6 & 2.2 \\
\textcolor{mygreen}{w/ \ssc{Few-shot}} & & 0.8 & 1.1 & 0.0 & 0.0 & 3.2 & 2.8 & 0.0 & 1.1 & 0.9 & 0.0 & 0.0 \\
\textcolor{mygreen}{w/ \ssc{CoT}} & & 2.2 & 2.9 & 0.8 & 0.8 & 7.1 & 7.1 & 0.4 & 3.4 & 1.9 & 0.0 & 0.0 \\
\bottomrule
\end{tabular}
\end{adjustbox}
\label{table:freshqa_relaxed_ood_llms}
\vspace{-3mm}
\end{table}

\clearpage
\newpage
\newpage
\begin{table}[t]
\centering
\caption{Accuracy of different search engine-augmented \llms on \freshqa under \relaxed evaluations. \hl{Models benchmarked on the same date of April 26, 2023}. We report accuracy across different categories of questions, including \textit{fast-changing} (\textit{fast}), \textit{slow-changing} (\textit{slow}), \textit{never-changing} (\textit{never}), false-premise, questions that involve knowledge before 2022 ($< 2022$) and since 2022 ($\geq 2022$), one-hop (\textit{1-hop}) and multi-hop (\textit{$m$-hop}) questions. $^{+}$ indicates a model with access to the current date. \ssc{UTD} stands for ``up-to-date''.}%
\begin{adjustbox}{max width=\textwidth}
\begin{tabular}{l c c c c c c c c c c c c c}
\toprule
\multirow{2}{*}{\textbf{Model}} & \multirow{2}{*}{\textbf{knowl.}} & \multirow{2}{*}{\textbf{all}} & \multicolumn{8}{c}{\textbf{valid premise}} & \multicolumn{2}{c}{\textbf{false premise}} \\
\cmidrule(l){4-11} \cmidrule(l){12-13} 
& \textbf{cutoff} & & {all} & {fast} & {slow} & {never} & {\mbox{\scriptsize $< 2022$}} & {\mbox{\scriptsize $\geq 2022$}} & {1-hop} & {$m$-hop} & {all} & {\mbox{\scriptsize $< 2022$}} \\
\multicolumn{13}{l}{\emph{comparison against baselines}} \\
\textcolor{mygreen}{\ssc{Google Search}} & UTD & 47.4 & 58.8 & 42.4 & 56.0 & 77.8 & 74.5 & 49.4 & 66.4 & 39.8 & 12.9 & 11.8 \\
\\
\textcolor{mygreen}{\gptthreefive} & 2021 & 32.4 & 32.4 & 8.0 & 28.0 & 61.1 & 68.1 & 11.1 & 34.7 & 26.9 & 32.3 & 43.0 \\
\textcolor{mygreen}{\gptthreefive + \ssc{Self-Ask}} & UTD & 42.0 & 51.6 & 36.8 & 44.8 & 73.0 & 74.5 & 37.9 & 53.0 & 48.1 & 12.9 & 17.2 \\
\textcolor{mygreen}{\gptthreefive + \seal} & UTD & 62.0 & 68.9 & 51.2 & 70.4 & 84.9 & 78.0 & 63.4 & 75.0 & 53.7 & 41.1 & 49.5 \\
\textcolor{mygreen}{\ssc{PPLX.AI}} & UTD & 66.2 & 68.9 & 48.8 & 67.2 & 90.5 & 85.1 & 59.1 & 76.1 & 50.9 & 58.1 & 60.2 \\
\\
\textcolor{mygreen}{\gptfour} & 2021$^{+}$ & 46.4 & 39.6 & 14.4 & 35.2 & 69.0 & 80.9 & 14.9 & 39.2 & 40.7 & 66.9 & 83.9 \\
\textcolor{mygreen}{\gptfour + \ssc{Self-Ask}} & UTD & 50.4 & 48.4 & 40.0 & 49.6 & 55.6 & 52.5 & 46.0 & 45.1 & 56.5 & 56.5 & 69.9 \\
\textcolor{mygreen}{\gptfour + \seal} & UTD & \textbf{77.8} & \textbf{78.7} & \textbf{61.6} & \textbf{79.2} & \textbf{95.2} & \textbf{90.8} & \textbf{71.5} & \textbf{83.2} & \textbf{67.6} & \textbf{75.0} & \textbf{80.6} \\
\midrule
\multicolumn{13}{l}{\emph{sensitivity and ablation studies}} \\
\textcolor{mygreen}{\gptthreefive} & 2021 & 32.4 & 32.4 & 8.0 & 28.0 & 61.1 & 68.1 & 11.1 & 34.7 & 26.9 & 32.3 & 43.0 \\
\textcolor{mygreen}{\gptthreefive + \seal} & UTD & 62.0 & 68.9 & 51.2 & 70.4 & 84.9 & 78.0 & 63.4 & 75.0 & 53.7 & 41.1 & 49.5 \\
\textcolor{mygreen}{w/ \ssc{premise check}} & UTD & 41.0 & 33.5 & 23.2 & 32.0 & 45.2 & 44.0 & 27.2 & 37.7 & 23.1 & 63.7 & 72.0 \\
\\
\textcolor{mygreen}{\gptfour} & 2021$^{+}$ & 46.4 & 39.6 & 14.4 & 35.2 & 69.0 & 80.9 & 14.9 & 39.2 & 40.7 & 66.9 & 83.9 \\
\\
\textcolor{mygreen}{\gptfour w/ \ssc{snippets only \& search order}} & UTD & 77.6 & 78.2 & 59.2 & \textbf{80.0} & 95.2 & 90.8 & 70.6 & 82.1 & 68.5 & 75.8 & 83.9 \\
\textcolor{mygreen}{\gptfour w/ \ssc{snippets only \& time order}} & UTD & 
77.6 & 78.2 & 59.2 & 79.2 & \textbf{96.0} & 90.1 & 71.1 & 82.1 & 68.5 & 75.8 & 86.0 \\
\textcolor{mygreen}{\gptfour w/ \ssc{snippets only \& random order}} & UTD & 75.4 & 76.1 & 58.4 & 73.6 & \textbf{96.0} & 90.8 & 67.2 & 80.6 & 64.8 & 73.4 & 81.7 \\
\\
\textcolor{mygreen}{\gptfour + \seal} & UTD & 77.8 & 78.7 & 61.6 & 79.2 & 95.2 & 90.8 & 71.5 & \textbf{83.2} & 67.6 & 75.0 & 80.6 \\
\textcolor{mygreen}{w/ \ssc{premise check}} & UTD & 78.8 & 76.3 & 59.2 & 76.8 & 92.9 & 87.2 & 69.8 & 82.1 & 62.0 & \textbf{86.3} & \textbf{90.3} \\
\textcolor{mygreen}{w/o \ssc{answer box}} & UTD & 76.2 & 76.6 & 59.2 & 76.0 & 94.4 & 90.1 & 68.5 & 81.0 & 65.7 & 75.0 & 80.6 \\
\textcolor{mygreen}{w/o \ssc{answer box} \& \ssc{relevant info}} & UTD & 74.8 & 75.0 & 56.0 & 74.4 & 94.4 & 89.4 & 66.4 & 80.6 & 61.1 & 74.2 & 81.7 \\
\textcolor{mygreen}{w/ \ssc{1 evidence}} & UTD & 67.2 & 67.3 & 47.2 & 66.4 & 88.1 & 85.8 & 56.2 & 72.0 & 55.6 & 66.9 & 79.6 \\
\textcolor{mygreen}{w/ \ssc{5 evidences}} & UTD & 74.2 & 75.0 & 56.8 & 74.4 & 93.7 & 87.2 & 67.7 & 81.7 & 58.3 & 71.8 & 77.4 \\
\textcolor{mygreen}{w/ \ssc{15 evidences}} & UTD & \textbf{79.0} & \textbf{79.5} & \textbf{62.4} & \textbf{80.0} & \textbf{96.0} & 90.1 & \textbf{73.2} & \textbf{83.2} & \textbf{70.4} & 77.4 & 81.7 \\
\textcolor{mygreen}{w/ \ssc{15 demonstrations}} & UTD & 77.2 & 78.2 & 60.0 & 78.4 & \textbf{96.0} & \textbf{91.5} & 70.2 & 82.8 & 66.7 & 74.2 & 79.6 \\
\textcolor{mygreen}{w/ \ssc{long demonstration answers}} & UTD & 77.8 & 77.9 & 60.8 & 77.6 & 95.2 & 90.1 & 70.6 & 82.8 & 65.7 & 77.4 & 83.9 \\
\bottomrule
\end{tabular}
\end{adjustbox}
\label{table:freshqa_relaxed_utd_llms}
\vspace{-3mm}
\end{table}

\appendix

\end{document}